\documentclass[10pt,twocolumn,letterpaper]{article}

\usepackage{wacv}              
\usepackage[accsupp]{axessibility}
\usepackage{graphicx}
\usepackage{amsmath}
\usepackage{amssymb}
\usepackage{booktabs}
\usepackage{arydshln}

\usepackage{float}
\usepackage{kotex}
\usepackage{multirow}
\usepackage{multicol}
\usepackage{colortbl}
\usepackage{subcaption}
\usepackage{makecell}
\usepackage{float}

\usepackage[sectionbib]{chapterbib}
\usepackage{diagbox}

\usepackage{enumitem}

\definecolor{light-gray}{gray}{0.82}

\usepackage{listings}
\usepackage{algorithm}
\usepackage[T1]{fontenc}
\usepackage[utf8]{inputenc} 
\definecolor{codeblue}{rgb}{0.25,0.5,0.5}
\definecolor{codekw}{rgb}{0.85, 0.18, 0.50}
\usepackage{highlight_table}
\newcommand{\g}[1]{\gradientcell{#1}{80}{100}{white}{gray}{70}}

\usepackage[pagebackref,breaklinks,colorlinks]{hyperref}

\usepackage[capitalize]{cleveref}
\crefname{section}{Sec.}{Secs.}
\Crefname{section}{Section}{Sections}
\Crefname{table}{Table}{Tables}
\crefname{table}{Tab.}{Tabs.}

\def\wacvPaperID{694} 
\def\confName{WACV}
\def\confYear{2025}

\begin{document}

\title{SFLD: Reducing the content bias for AI-generated Image Detection}
\author{\textbf{Seoyeon Gye\thanks{Equal contribution.}\hspace{0.6cm} Junwon Ko\footnotemark[1]\hspace{0.6cm} Hyounguk Shon\footnotemark[1]\hspace{0.6cm} Minchan Kwon\hspace{0.7cm} Junmo Kim}\vspace{0.3cm} \\
School of Electrical Engineering, KAIST, South Korea\\
{\tt\small \{sawyun, kojunewon, hyounguk.shon, kmc0207, junmo.kim\}@kaist.ac.kr}}
\maketitle

\begin{abstract}
    Identifying AI-generated content is critical for the safe and ethical use of generative AI.
    Recent research has focused on developing detectors that generalize to unknown generators, with popular methods relying either on high-level features or low-level fingerprints.
    However, these methods have clear limitations: biased towards unseen content, or vulnerable to common image degradations, such as JPEG compression.
    To address these issues, we propose a novel approach, SFLD, which incorporates PatchShuffle to integrate high-level semantic and low-level textural information. SFLD applies PatchShuffle at multiple levels, improving robustness and generalization across various generative models.
    Additionally, current benchmarks face challenges such as low image quality, insufficient content preservation, and limited class diversity. In response, we introduce TwinSynths, a new benchmark generation methodology that constructs visually near-identical pairs of real and synthetic images to ensure high quality and content preservation.
    Our extensive experiments and analysis show that SFLD outperforms existing methods on detecting a wide variety of fake images sourced from GANs, diffusion models, and TwinSynths, demonstrating the state-of-the-art performance and generalization capabilities to novel generative models.
    The TwinSynths dataset is publicly available at \textnormal{\url{https://huggingface.co/datasets/koooooooook/TwinSynths}}.
\end{abstract}

\section{Introduction}
\label{sec:intro}

\begin{figure}[t]
    \centering
    \begin{subfigure}[t]{0.48\columnwidth}
        \centering
        \includegraphics[width=\textwidth]{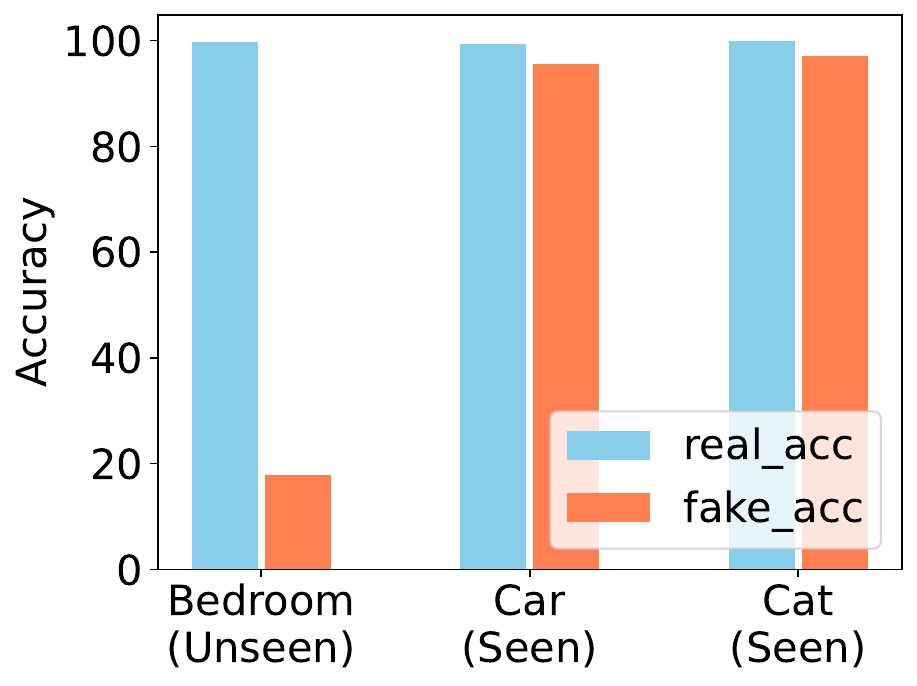}
        \caption{UnivFD\cite{ojha2023towards}}
        \label{fig:univfd}
    \end{subfigure}
    \begin{subfigure}[t]{0.48\columnwidth}
        \centering
        \includegraphics[width=\textwidth]{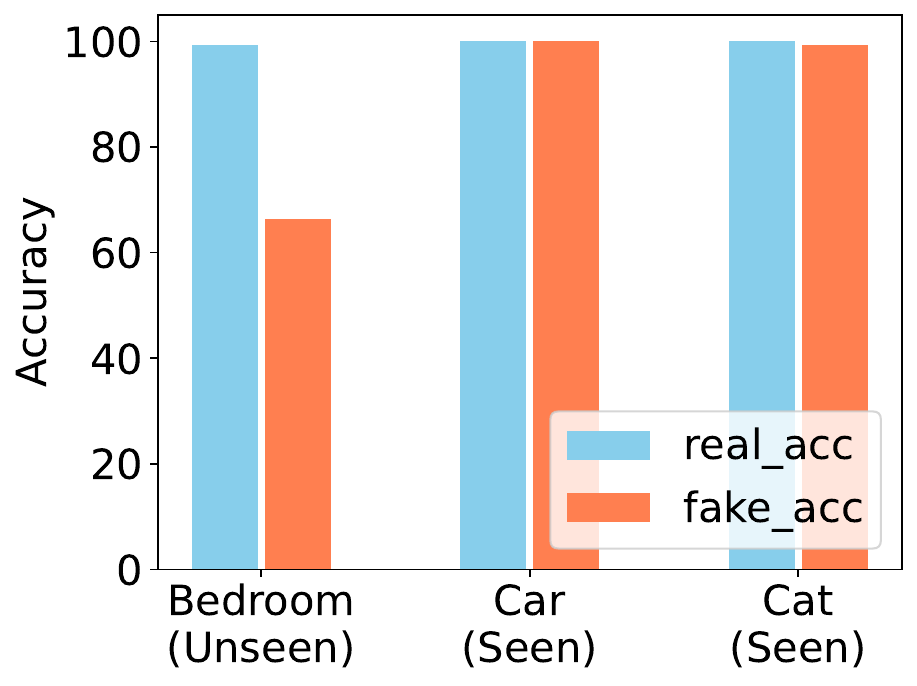}
        \caption{SFLD (ours)}
        \label{fig:ours}
    \end{subfigure}
    \caption{Class-wise detection accuracy for StyleGAN-\{\textit{bedroom, car, cat}\} class categories. The \textit{bedroom} class does not appear at training, while \textit{car} and \textit{cat} does. UnivFD\cite{ojha2023towards} catastrophically fails to identify synthetic bedroom images, which hints at model bias towards high-level image content.}
    \label{fig:overview}
\end{figure}

\begin{figure*}[t]
    \centering
    \begin{subfigure}[t]{0.45\linewidth}
        \centering
        \includegraphics[width=\linewidth]{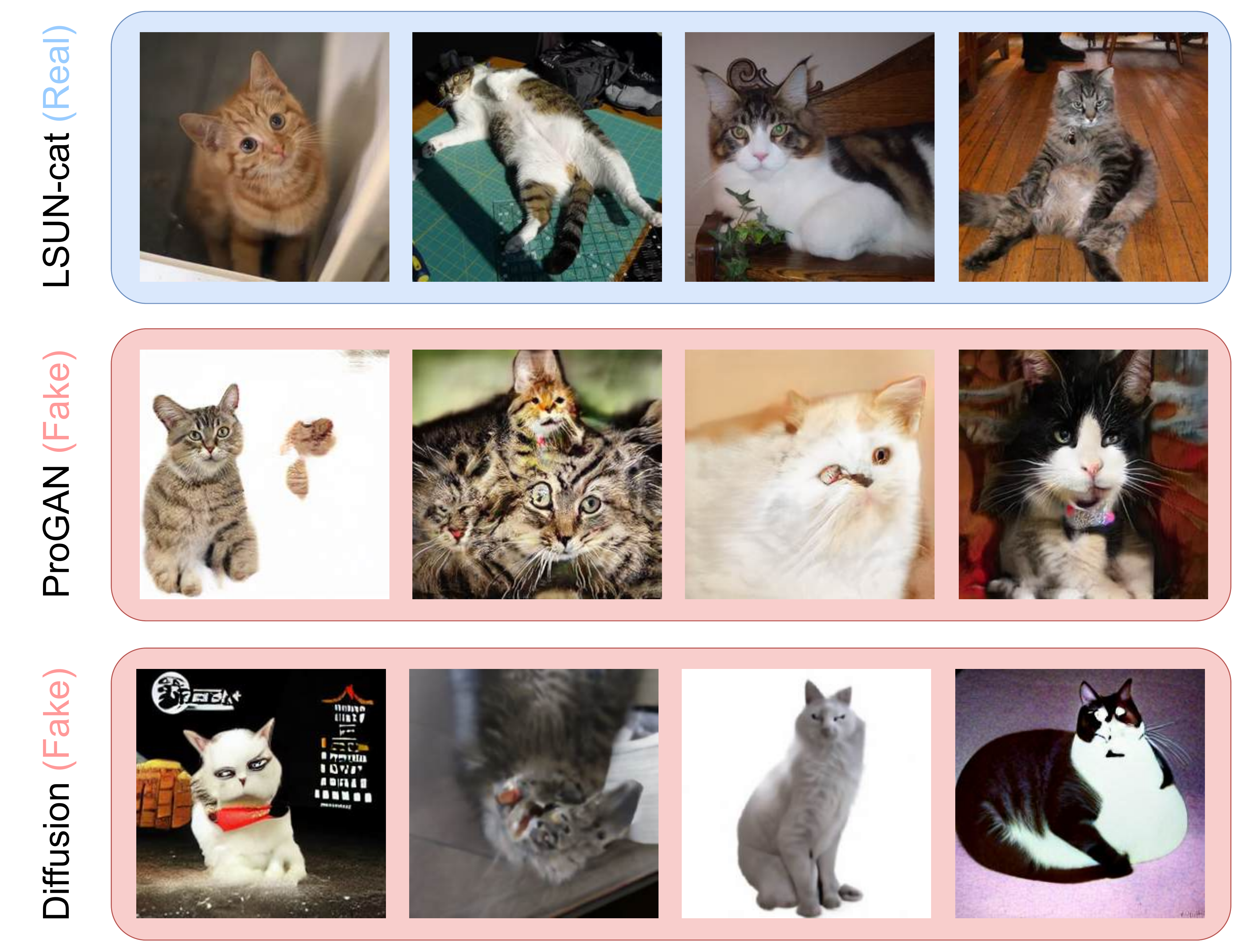}
        \caption{Examples of the conventional benchmark.}
        \label{fig:benchmark_visual_trad}
    \end{subfigure}
    \begin{subfigure}[t]{0.45\linewidth}
        \centering
        \includegraphics[width=\linewidth]{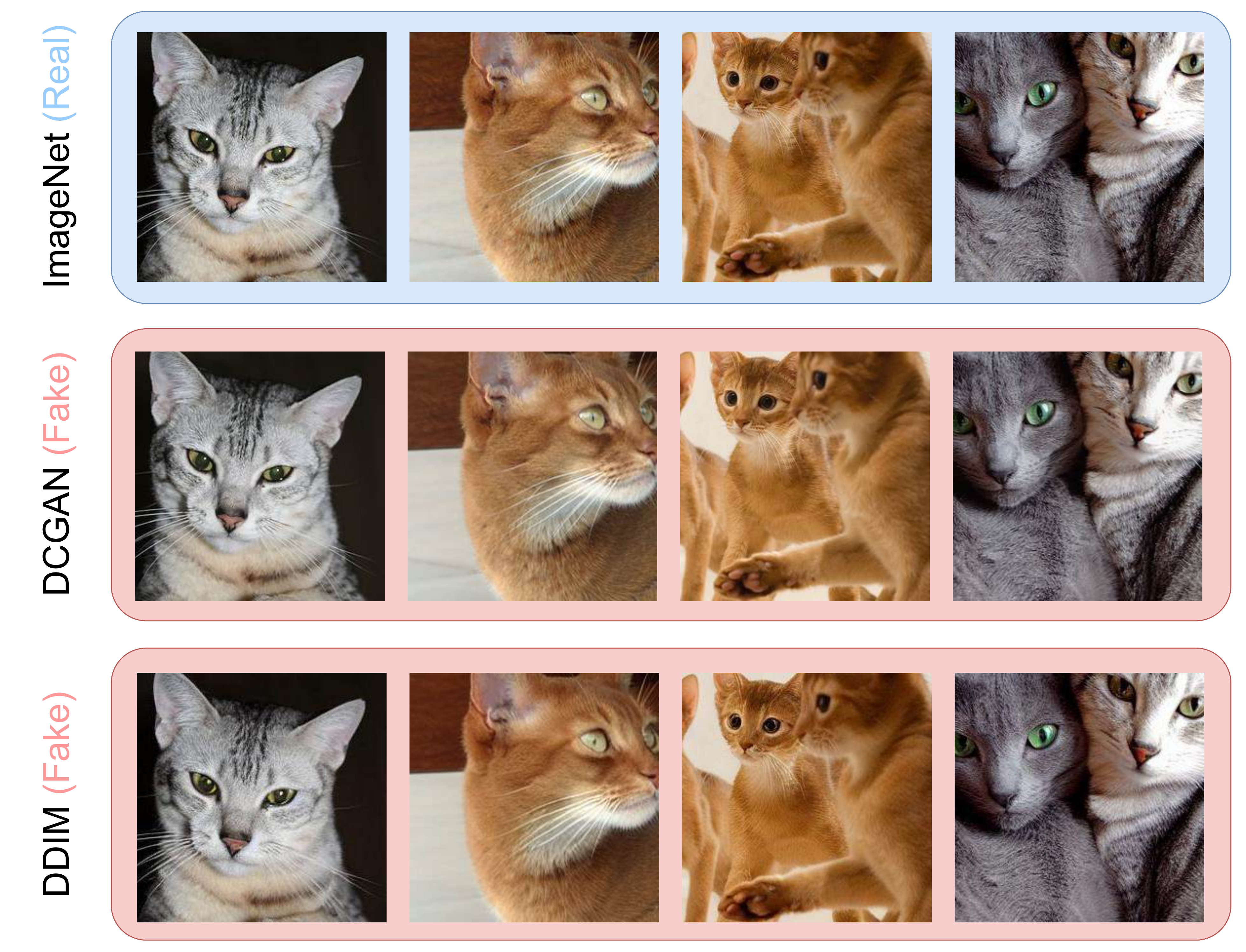}
        \caption{Examples of proposed TwinSynths benchmark.}
        \label{fig:benchmark_visual_new}
    \end{subfigure}
    
    \caption{Comparison of benchmarks. (a) Real images and fake GAN images are sampled from the test ProGAN set in the ForenSynths\cite{wang2020cnn}. Fake diffusion images are sampled from benchmark of Ojha \etal\cite{ojha2023towards}, each from LDM, GLIDE and DALL-E dataset. (b) Real images are sampled from ImageNet dataset, and corresponding fake images are generated by each model.}
    \label{fig:benchmark_visual}
\end{figure*}

The rapid advancement of AI image generation technologies has brought significant achievements but also growing social concern, as these technologies are increasingly misused for the creation of fake news, malicious defamation, and other forms of digital deception. 
In response, AI-generated image detection is receiving more attention.
There is a wide variety of generative models, along with commercial models with unknown internal architectures.
This highlights the need for a generalized detector capable of distinguishing between real and fake images, regardless of the generative model structure.

In this context, early research focused on identifying the characteristic fingerprints of generated images.
Recent work, NPR\cite{tan2024rethinking} shows that pixel-level features, induced by the upsampling layers commonly found in current generative models, can serve as cues for detection.
However, there are clear practical limitations to relying on low-level fingerprints. 
First, the approach is vulnerable to simple image degradations, such as JPEG compression or blurring, which are common in real-world online environments\cite{wang2020cnn}.
Additionally, the model may become biased toward the specific \emph{fakeness} seen at training in cases where generalization to novel generators is not sufficiently considered\cite{ojha2023towards, zhu2023gendet}.
For instance, a detector trained on GAN-generated images may learn the characteristics of GANs as the fake features, while mistakenly perceiving images generated by diffusion models as real.
This bias limits the detector's generalizability across different types of generative models.

To tackle these limitations, UnivFD\cite{ojha2023towards} utilizes a robust, pre-trained image encoder.
This image embedding is task-agnostic, enabling it to capture high-level semantic information from images.
However, we found that UnivFD exhibits a bias towards the observed content in the training images, learning another specific \textit{fakeness}.
\cref{fig:overview} shows that UnivFD misclassifies most GAN-generated images of a novel class (StyleGAN-\textit{bedroom}) as \textit{real}.
The \textit{bedroom} class is absent from the training set, which may lead the detector to mistakenly classify most images as real, demonstrating the detector's reliance on seen content during training.

We propose a novel technique called \textbf{PatchShuffle}, which is the core of our fake image detection model, \textbf{SFLD} (pronounced ``shuffled'').
PatchShuffle divides the image into non-overlapping patches and randomly shuffles them.
This procedure disrupts the high-level semantic structure of the image while preserving low-level textural information. 
This allows the detection model to better focus on both context and texture.
SFLD utilizes an ensemble of classifiers at multiple levels of PatchShuffle, leveraging hierarchical information across various patch sizes.
This approach ensures that the model leverages both the semantic and textural aspects of the image to improve fake image detection.
The results demonstrate that SFLD achieves superior performance with enhanced robustness and better generalization.

Furthermore, we observe that previous benchmarks have three limitations:
\textbf{(1) low image quality.}
The previous benchmarks contain a significant portion of low-quality images that lag behind the capabilities of current generative models.
As a result, the practical usefulness of these benchmarks is significantly reduced.
\textbf{(2) lack of content preservation.}
Some subsets---particularly foundation generative models---lack access to the training data used for the checkpoints.
Consequently, the content of the generated and real images often differs significantly, making it difficult to determine whether a detector focuses on real/fake discriminative features or other irrelevant features. 
\textbf{(3) limited class diversity.}
Existing benchmarks primarily focus on expanding the variety of generative models without considering the generated class diversity and scalability among generative models.
As shown in \cref{fig:overview}, this makes it difficult to identify detection bias towards certain classes, as well as hard to represent the in-the-wild performance of the detector due to limited class diversity.

To address these challenges, we propose a new benchmark generation methodology and corresponding benchmark, \textbf{TwinSynths}.
It consists of synthetic images that are visually near-identical to paired real images for practical and fair evaluations.
TwinSynths constructs image pairs that preserve both quality and content while retaining the architectural characteristics of each generative model.
Also, TwinSynths enables flexible class expansion by generating synthetic images tailored to the real image. 
Using this benchmark, we evaluate the performance of our proposed SFLD method as well as existing detection models.

Our main contributions are summarized as follows: 
\begin{itemize}[leftmargin=*]
    \setlength\itemsep{0.0em}
    \item We propose SFLD, a novel AI-generated image detection method that integrates semantic and texture artifacts on generated images, achieving state-of-the-art performance.
    \item We propose a new approach on benchmarks and the subset of generated images that can ensure the quality and content of generated images.
    \item We validate our method through extensive experiments and analysis that support our hypothesis.
\end{itemize}

\section{Method}
\label{sec:method}

\begin{figure}[t]
     \centering
     \includegraphics[width=1.0\linewidth]{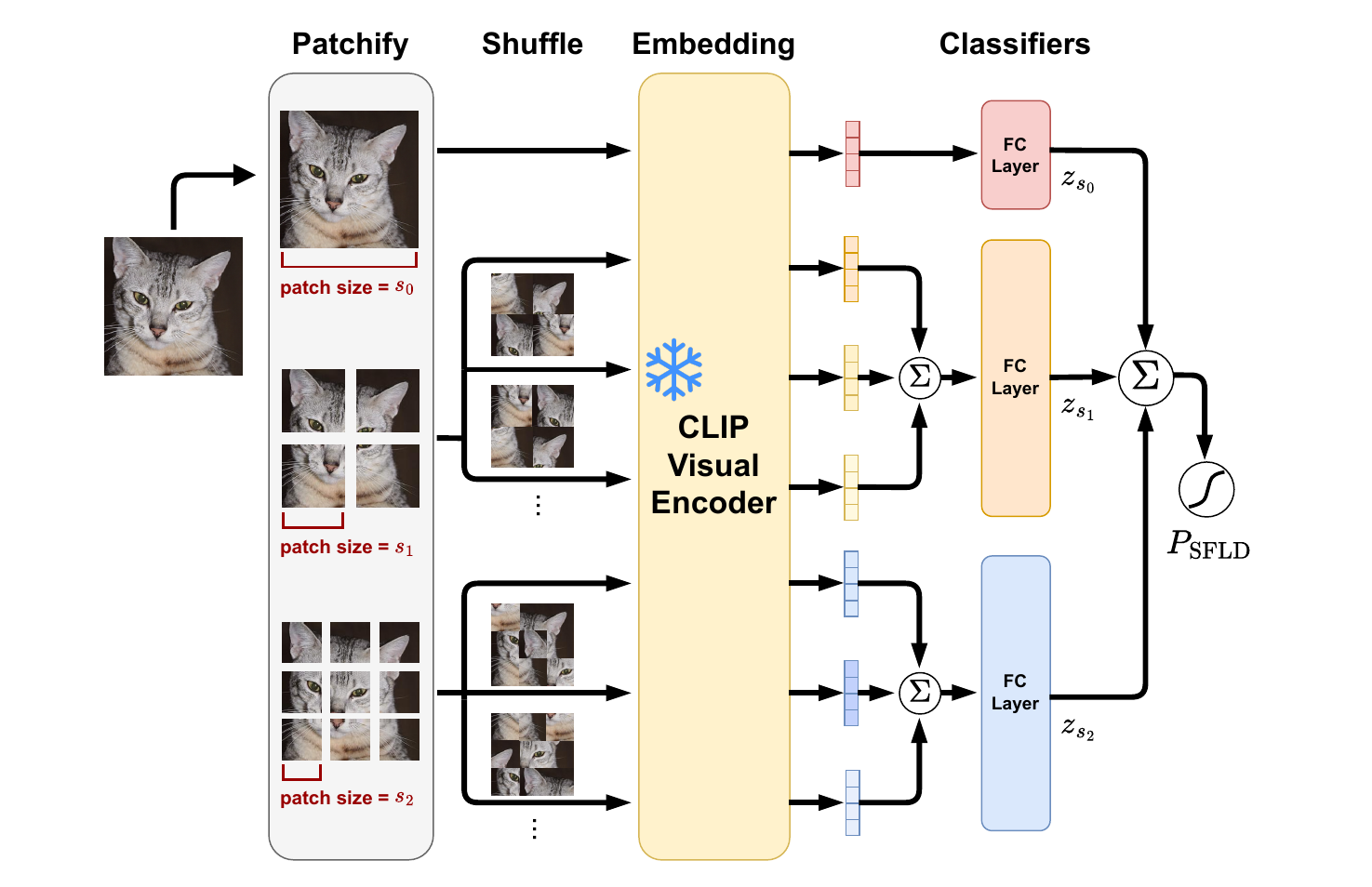}
     \caption{Architecture of the proposed fake image detector (SFLD). $z_{s_i}$ refers to the logit score generated from an input image processed via $s_i\text{×}s_i$ patch size. $\Sigma$ indicates weighted sum.}
     \label{fig:method-architecture}
\end{figure}

\subsection{Patch Shuffling Fake Detection} 
\label{subsec:SFLD}
\textbf{Backbone.}
We utilize the visual encoder of CLIP ViT-L/14\cite{dosovitskiy2021an, CLIP} to leverage the pre-trained feature space. This choice is based on Ojha \etal\cite{ojha2023towards}, which showed that it outperforms other models such as CLIP:ResNet-50, ImageNet:ResNet-50, and ImageNet:ViT-B/16 in distinguishing real from fake images. The results indicated that both the architecture and the pre-training data are crucial. 
Based on this insight, we chose the ViT model for our backbone. 
As shown in \cref{fig:method-architecture}, we extract CLIP features and train a fully connected layer to classify real and fake images.

\textbf{PatchShuffle.}
To effectively integrate both semantic and textural features, PatchShuffle disrupts the global structure of an image while preserving local features.
In the PatchShuffle process, the input images are divided into non-overlapping patches of size $s \times s$ and then randomly shuffled. This operation produces a new shuffled image $x_s$.

For a given $s$, the logit score of the shuffled image is, 
\begin{equation}
    z_{s} = \psi(f(x_s)) \,,
\end{equation}
where $f( \cdot )$ represents a pre-trained CLIP encoder and $\psi(\cdot)$ is a single fully connected layer appended to $f$.

There are classifiers for each patch size of shuffled images to leverage local structure information hierarchically within the image.
We selected patch sizes of 28, 56, and 224 for the proposed SFLD.
As shown in \cref{fig:method-architecture}, $s_0$ is 224, $s_1$ is 56 and $s_2$ is 28.
These configurations are studied in detail in \cref{subsec:patchsetting}.
For each patch size $s_j$, the classifier $\psi_{s_j}$ is trained independently.
Notably, UnivFD takes a center-cropped 224×224 image as input to the CLIP encoder.
Therefore, when using a patch size of 224 in PatchShuffle, it effectively corresponds to the same setting as UnivFD\cite{ojha2023towards}.

We employ binary cross-entropy loss for each classifier:
\small\begin{equation}
    \mathcal{L} = -\frac{1}{N} \sum_{i=1}^{N} \left[ y_i \log \sigma(z_{s_j}) + (1 - y_i) \log \left(1 - \sigma(z_{s_j}) \right) \right] 
\end{equation}\normalsize
where $N$ is the number of data and $y_i \in \{0, 1\}$ is the label whether an input $x_i$ is real ($y_i = 0$) or fake ($y_i = 1$).

\textbf{SFLD.} SFLD combines multiple classifiers trained on shuffled images with different patch sizes.
By varying the patch size, SFLD incorporates models that focus on various levels of structural features, ranging from fine-grained local details to more global patterns.

During testing, $N_{views}=10$ shuffled views are generated for each patch size. The logits from these views are averaged and processed by the corresponding classifier. The final probability $P_{SFLD}(y|x)$ is computed by averaging the logits across patch sizes and applying the sigmoid function:
\small\begin{equation}
    P_{\text{SFLD}}(y|x) = \sigma\left(\frac{1}{k} \sum_{j=1}^{k} \psi_{s_j}\left(\frac{1}{N_{\text{views}}} \sum_{i=1}^{N_{\text{views}}} f(x_{s_j}^i)\right)\right) \,,
\end{equation}\normalsize
where $k$ is the number of patch sizes used in the ensemble (e.g., $k=3$ in our configuration).

Binary classification is done using a threshold of 0.5 on $P_{SFLD}$. Although the fusion method is simple and not tuned for each test class, its simplicity enables strong generalization across diverse fake image sources. By combining classifiers trained on different patch sizes, SFLD achieves a robust and general detection performance. \cref{alg:pseudocode} shows the full workflow of SFLD, especially the fusion of multiple classifiers during inference.

\subsection{TwinSynths} 
In \cref{sec:intro}, we pointed out three shortcomings in the previous benchmarks: low image quality, lack of content preservation, and limited class diversity.
This issue must be addressed to allow a comprehensive comparison of detectors.
Therefore, we propose a novel dataset creation methodology and \emph{TwinSynths} benchmark, consisting of GAN- and diffusion-based generated images that are  paired with visually-identical real counterparts.
To create a practical benchmark for evaluating generated image detectors, it is essential to ensure the generation of high-quality images that preserve the original content.
To achieve this, the image generation process should ideally sample a distribution that closely resembles a real distribution.
From this perspective, the image generation or sampling process can be interpreted as effectively fitting the generator to a single real image.
Through this approach, we construct image pairs that preserve the content of the images while reflecting the architectural traits of the generative models.
Additionally, this methodology allows for the expansion of target classes in the benchmark by generating paired images for any real image.
\cref{fig:benchmark_visual_new} are some examples of TwinSynths.
We can see that the content of the paired real image is faithfully reproduced and the quality of the generated image is guaranteed.

\textbf{TwinSynths-GAN benchmark.}
The GAN-based subsets in the previous benchmark have disparate training configurations, especially the class of training images, resulting in a discrepancy between the generated and the real images.
In order to generate a high quality image that preserves the content of the paired real image while leveraging the training methodology of GANs, we trained the generator from scratch using a single real image.
The MSE loss was provided to the generator to generate an image that is identical to the original image.
For reproduction, the latent vector for the generator input is maintained at a fixed value.
We created 8,000 generated images from 80 selected ImageNet\cite{russakovsky2015imagenet} classes, which is much larger than previous benchmarks.
We selected 40 classes following the \emph{ProGAN} subset in ForenSynths\cite{wang2020cnn}, while the other 40 classes were chosen arbitrarily. 
We utilized DCGAN \cite{radford2015unsupervised} architecture. 

\textbf{TwinSynths-DM benchmark.} 
In comparison to GAN-based subsets, diffusion-based subsets in conventional benchmarks were generated with off-the-shelf pretrained models, having much severer content discrepancy between real and generated images.
In order to generate a high quality image that preserves the content of paired real image while leveraging the inference process of diffusion models, we used DDIM inversion\cite{songdenoising} to generate image that is similar to the real image.
We apply a DDIM forward process to the real image to make it noisy and perform text-conditioned DDIM denoising process using the prompt template \texttt{`a photo of  \{class name\}'}. 
For the prompts, we used the class names from ImageNet.
This process makes TwinSynths-DM preserve the similarity with the paired real images. 
We used the same image classes used to create TwinSynths-GAN. 
We utilized the pretrained decoder and scheduler of \cite{songdenoising}.

\section{Experiments}
\label{sec:experiments}

\begin{table*}[ht]
    \centering
    \scriptsize
    \setlength{\tabcolsep}{2pt}
    \resizebox{\textwidth}{!}{
        \renewcommand{\g}[1]{\gradientcell{#1}{88}{100}{white}{gray}{70}}

\begin{tabular}{lcccccccccccccccc|c}
    \toprule
    Method & \makecell{Pro\\GAN} & \makecell{Style\\GAN} & \makecell{Style\\GAN2} & \makecell{Big\\GAN} & \makecell{Cycle\\GAN} & \makecell{Star\\GAN} & \makecell{Gau\\GAN} & \makecell{Deep\\fake} & \makecell{DALL\\E} & \makecell{Glide\\100\_10} & \makecell{Glide\\100\_27} & \makecell{Glide\\50\_27} & ADM & \makecell{LDM\\100} & \makecell{LDM\\200} & \makecell{LDM\\200\_cfg} & Avg. \\ 
    \midrule
    CNNSpot\cite{wang2020cnn} & \g{100} & \g{99.8} & \g{99.5} & \g{86.0} & \g{94.9} & \g{99.0} & \g{90.8} & \g{84.5} & \g{72.9} & \g{82.5} & \g{80.1} & \g{84.7} & \g{78.3} & \g{71.5} & \g{70.3} & \g{73.6} & \g{85.53} \\
    FreDect\cite{Frank} & \g{100} & \g{96.3} & \g{72.7} & \g{93.9} & \g{88.8} & \g{99.4} & \g{84.5} & \g{71.9} & \g{95.0} & \g{52.2} & \g{53.9} & \g{55.0} & \g{57.3} & \g{93.1} & \g{92.7} & \g{90.4} & \g{81.07} \\
    GramNet\cite{liu2020global} & \g{100} & \g{88.2} & \g{100} & \g{62.7} & \g{74.2} & \g{100} & \g{55.0} & \g{93.5} & \g{98.8} & \g{99.7} & \g{99.3} & \g{99.1} & \g{79.8} & \g{99.8} & \g{99.8} & \g{99.8} & \g{90.61} \\
    Fusing\cite{ju2022fusing} & \g{100} & \g{97.5} & \g{100} & \g{89.1} & \g{95.5} & \g{99.8} & \g{87.7} & \g{69.3} & \g{77.1} & \g{83.6} & \g{81.3} & \g{86.2} & \g{82.6} & \g{75.5} & \g{76.2} & \g{77.9} & \g{86.20} \\
    LNP\cite{liu2022detecting} & \g{100} & \g{92.5} & \g{100} & \g{90.2} & \g{93.9} & \g{100} & \g{77.9} & \g{73.7} & \g{94.9} & \g{92.1} & \g{88.5} & \g{89.5} & \g{85.5} & \g{93.9} & \g{93.6} & \g{93.7} &  \g{91.24} \\
    LGrad\cite{Tan2023CVPR} & \g{100} & \g{84.2} & \g{99.9} & \g{87.9} & \g{94.4} & \g{100} & \g{91.7} & \g{64.3} & \g{95.6} & \g{97.1} & \g{94.8} & \g{96.3} & \g{74.9} & \g{96.3} & \g{96.2} & \g{96.5} & \g{91.88}  \\ 
    UnivFD\cite{ojha2023towards} & \g{100} & \g{97.2} & \g{98.0} & \g{99.3} & \g{99.8} & \g{99.4} & \g{100} & \g{81.8} & \g{97.7} & \g{95.5} & \g{95.8} & \g{96.0} & \g{88.3} & \g{99.4} & \g{99.4} & \g{93.2} & \g{96.29} \\
    NPR\cite{tan2024rethinking} & \g{100} & \g{99.4} & \g{99.9} & \g{87.4} & \g{90.0} & \g{100} & \g{76.7} & \g{82.7} & \g{99.2} & \g{100} & \g{99.8} & \g{99.9} & \g{84.2} & \g{100} & \g{99.9} & \g{99.9} & \g{94.94} \\ 
    \midrule
    SFLD (224+28) & \g{100} & \g{99.8} & \g{99.9} & \g{99.9} & \g{100} & \g{100} & \g{100} & \g{91.5} & \g{99.1} & \g{96.7} & \g{97.0} & \g{97.5} & \g{94.5} & \g{99.3} & \g{99.3} & \g{94.2} & \g{98.03} \\ 
    SFLD (224+56) & \g{100} & \g{99.8} & \g{99.9} & \g{99.8} & \g{100} & \g{100} & \g{100} & \g{90.9} & \g{99.2} & \g{98.2} & \g{98.4} & \g{98.7} & \g{94.4} & \g{99.6} & \g{99.6} & \g{95.8} & \g{98.39}\\ 
    SFLD & \g{100} & \g{99.9} & \g{99.9} & \g{99.9} & \g{100} & \g{100} & \g{100} & \g{93.3} & \g{99.3} & \g{97.6} & \g{97.9} & \g{98.4} & \g{95.4} & \g{99.3} & \g{99.3} & \g{95.0} & \g{98.43}\\ 
    \bottomrule
\end{tabular}

    }
    \caption{Generalization performance on the conventional benchmark reported in AP. SFLD (224+28) indicates the ensemble of the classifier with patch sizes 224 and 28. And SFLD indicates the ensemble of the three classifiers with patch sizes 224, 56, and 28.}
    \label{tab:main-results-map}
\end{table*}

\begin{table*}[ht]
    \centering
    \scriptsize
    \setlength{\tabcolsep}{2pt}
    \resizebox{\textwidth}{!}{
        \renewcommand{\g}[1]{\gradientcell{#1}{75}{100}{white}{gray}{70}}

\begin{tabular}{lcccccccccccccccc|c}
    \toprule
    Method & \makecell{Pro\\GAN} & \makecell{Style\\GAN} & \makecell{Style\\GAN2} & \makecell{Big\\GAN} & \makecell{Cycle\\GAN} & \makecell{Star\\GAN} & \makecell{Gau\\GAN} & \makecell{Deep\\fake} & \makecell{DALL\\E} & \makecell{Glide\\100\_10} & \makecell{Glide\\100\_27} & \makecell{Glide\\50\_27} & ADM & \makecell{LDM\\100} & \makecell{LDM\\200} & \makecell{LDM\\200\_cfg} & Avg. \\ 
    \midrule
    CNNSpot\cite{wang2020cnn}& \g{100} & \g{90.2} & \g{86.9} & \g{71.2} & \g{87.6} & \g{94.6} & \g{81.4} & \g{50.7} & \g{57.7} & \g{62.4} & \g{61.3} & \g{64.4} & \g{62.5} & \g{54.9} & \g{54.8} & \g{56.0} & \g{71.02} \\
    FreDect\cite{Frank}  & \g{99.4} & \g{80.3} & \g{56.1} & \g{82.7} & \g{81.6} & \g{94.5} & \g{81.0} & \g{62.5} & \g{81.6} & \g{49.7} & \g{52.2} & \g{53.4} & \g{57.8} & \g{79.3} & \g{79.0} & \g{76.7} & \g{72.97} \\ 
    GramNet\cite{liu2020global} & \g{100} & \g{50.8} & \g{100} & \g{67.9} & \g{72.8} & \g{100} & \g{57.4} & \g{62.0} & \g{87.8} & \g{95.6} & \g{93.4} & \g{91.8} & \g{79.5} & \g{98.7} & \g{98.5} & \g{98.5} & \g{84.65} \\ 
    Fusing\cite{ju2022fusing} & \g{100} & \g{71.0} & \g{97.1} & \g{76.7} & \g{85.7} & \g{97.2} & \g{76.1} & \g{53.0} & \g{56.1} & \g{60.9} & \g{59.7} & \g{61.6} & \g{62.4} & \g{53.8} & \g{54.5} & \g{56.0} & \g{70.10} \\ 
    LNP\cite{liu2022detecting} & \g{99.8} & \g{78.1} & \g{99.6} & \g{81.1} & \g{82.1} & \g{99.9} & \g{71.7} & \g{56.1} & \g{83.5} & \g{80.3} & \g{76.7} & \g{78.0} & \g{67.2} & \g{80.6} & \g{79.6} & \g{81.7} & \g{80.98}  \\ 
    LGrad\cite{Tan2023CVPR} & \g{99.7} & \g{71.4} & \g{96.0} & \g{80.3} & \g{86.6} & \g{98.4} & \g{80.3} & \g{51.9} & \g{86.0} & \g{90.4} & \g{87.1} & \g{90.0} & \g{68.1} & \g{87.9} & \g{87.4} & \g{87.8} & \g{84.30}  \\ 
    UnivFD\cite{ojha2023towards}  & \g{100} & \g{84.4} & \g{75.7} & \g{95.2} & \g{98.7} & \g{95.9} & \g{99.7} & \g{67.7} & \g{87.5} & \g{78.1} & \g{78.7} & \g{79.2} & \g{70.0} & \g{95.2} & \g{94.6} & \g{74.2} & \g{85.89} \\ 
    NPR\cite{tan2024rethinking} & \g{100} & \g{95.4} & \g{96.9} & \g{82.9} & \g{90.0} & \g{99.9} & \g{79.8} & \g{74.6} & \g{83.0} & \g{97.9} & \g{96.6} & \g{97.1} & \g{74.3} & \g{98.0} & \g{97.9} & \g{97.7} & \g{91.38} \\ 
    \midrule
    SFLD (224+28) & \g{100} & \g{95.8} & \g{89.0} & \g{97.2} & \g{99.1} & \g{99.3} & \g{97.8} & \g{80.1} & \g{94.6} & \g{87.0} & \g{87.1} & \g{88.9} & \g{83.9} & \g{95.6} & \g{95.5} & \g{80.8} & \g{91.94} \\ 
    SFLD (224+56) & \g{100} & \g{90.6} & \g{86.5} & \g{97.8} & \g{99.5} & \g{99.0} & \g{98.9} & \g{82.7} & \g{94.0} & \g{89.2} & \g{89.2} & \g{90.9} & \g{81.0} & \g{97.0} & \g{96.6} & \g{80.1} & \g{92.05}\\ 
    SFLD & \g{100} & \g{96.7} & \g{91.9} & \g{96.5} & \g{99.2} & \g{99.4} & \g{96.0} & \g{84.2} & \g{95.2} & \g{90.6} & \g{90.7} & \g{92.5} & \g{86.0} & \g{95.6} & \g{95.7} & \g{82.9} & \g{93.30}\\ 
    \bottomrule
\end{tabular}

    }
    \caption{Generalization performance on the conventional benchmark reported in accuracy.}
    \label{tab:main-results-acc}
\end{table*}

\subsection{Settings}
\label{subsec:settings}

\textbf{Datasets}. 
Following the conventions of AI-generated image detection, all detectors were trained using the ForenSynths train set\cite{wang2020cnn}. This train set consists of real images used to train ProGAN\cite{karras2018progressive} and ProGAN-generated images. We evaluate the performance of SFLD on several benchmarks, including conventional benchmarks, TwinSynths, and low-level vision/perceptual loss benchmarks. For more detailed descriptions of the datasets and configurations used, please refer to \cref{supp:datasets}.

\textbf{Baseline methods}. 
We compare the performance of the proposed SFLD with existing AI-generated image detection methods. It includes CNNSpot\cite{wang2020cnn}, FreDect\cite{Frank}, GramNet\cite{liu2020global}, Fusing\cite{ju2022fusing}, LNP\cite{liu2022detecting}, LGrad\cite{Tan2023CVPR}, UnivFD\cite{ojha2023towards}, and NPR\cite{tan2024rethinking}. 
We conducted evaluations on the detection methods with our test dataset. 
The evaluation is done by the official models\cite{wang2020cnn, ojha2023towards}, re-implemented models\cite{Frank, liu2020global, ju2022fusing, liu2022detecting, Tan2023CVPR} by Zhong \etal\cite{rptc-AIGCDetection}, or trained model with the official codes using 20-classes train set \cite{tan2024rethinking}.

\textbf{Evaluation metrics}. 
We assess the performances of the detection models by average precision score (AP) and classification accuracy (Acc.), following previous works\cite{wang2020cnn, ojha2023towards, tan2024rethinking}. The AP metric is not dependent on the threshold value, whereas the Acc. is calculated with a fixed threshold of 0.5 across all generation models.

\subsection{Results on Conventional Benchmark}
\cref{tab:main-results-map,tab:main-results-acc} shows the detection performance on conventional benchmarks in AP and Acc.
All baselines are trained on only the ProGAN train dataset consisting of 20 classes.
Higher performance is colored darker.
SFLD demonstrates robust and generalized performance across various generators in the benchmark. 
Note that SFLD achieves above 90.0 AP on every unseen generator. 
SFLD has an average of 98.43 AP, outperforming the best-performing baseline, UnivFD, by up to 2.14 in average. 
While for some tasks NPR has shown outperforming AP values in some generators, it has shown relatively low performance on some generators.
In this regard, we found that NPR is sensitive to some image degradation or different post-processing methods in different generative models, which limits its practicality.
Refer to \cref{sec:discussion_robustness} for further comparison of robustness on image degradation.

SFLD also exhibits state-of-the-art performance in classification accuracy. It performs particularly well on challenging datasets like DeepFake and ADM. On DeepFake, it improves accuracy from 74.6\% to 84.2\% (+9.6), and on ADM, from 79.5\% to 86.0\% (+6.5). These gains highlight its superior generalization in difficult scenarios.

\subsection{Analysis on TwinSynths} 
\label{sec:experiments_bench_ours}

\begin{table}[t]
    \centering
    \resizebox{0.8\linewidth}{!}{
        \renewcommand{\g}[1]{\gradientcell{#1}{50}{80}{white}{gray}{70}}

\begin{tabular}{lccc}
\toprule
Method & Twin-GAN & Twin-DM & Avg. \\ 
  \midrule
CNNSpot    & \g{62.92} & \g{46.93} & \g{54.93} \\
FreDect    & \g{54.57} & \g{55.64} & \g{55.11} \\
GramNet       & \g{71.98} & \g{36.10} & \g{54.04} \\
Fusing     & \g{61.80} & \g{48.62} & \g{55.21} \\
LGrad      & \g{59.51} & \g{34.25} & \g{46.88} \\
UnivFD\cite{ojha2023towards}    & \g{58.09} & \g{74.38} & \g{66.24} \\
NPR\cite{tan2024rethinking}        & \g{78.19} & \g{35.76} & \g{56.98} \\ 
\midrule
PatchShuffle (28) & \g{73.56} & \g{65.52} & \g{69.54} \\
PatchShuffle (56) & \g{75.90} & \g{60.73} & \g{68.32} \\
SFLD (224+28) & \g{70.43} & \g{75.80} & \g{73.12} \\
SFLD (224+56) & \g{70.16} & \g{72.44} & \g{71.30} \\
SFLD & \g{73.82} & \g{72.05} & \g{72.94} \\ 
\bottomrule
\end{tabular}

    }
    \caption{Performance comparisons on TwinSynths. Values indicate AP score. \emph{DM} refers to diffusion model.}
    \label{tab:bench_ours}
\end{table}

\cref{tab:bench_ours} illustrates the detection performance on TwinSynths in AP.
The results demonstrate that SFLD is effective in TwinSynths while some detectors have shown a significant drop in performance.
Note that the TwinSynths focused on three key aspects: image quality, content preservation, and class diversity.
This suggests that the high performance on conventional benchmarks may not guarantee the detector's performance in real-world scenarios.

The results of TwinSynths allow an indirect analysis of the factors that the detectors focus on.
For convenience, we now define high-level features and low-level features.
high-level features are semantic information and their artifacts originate from distribution disparity between real images and generated images.
low-level features are texture information and their artifacts stem from the generator traces and image quality of generated images.
The TwinSynths-GAN preserves the content of the real image with minimal alteration, as the images are generated from a single real image.
This results in UnivFD, which captures high-level feature artifacts on the entire image, resulting in poor performance on the TwinSynths-GAN subset.
In contrast, NPR, which captures high-frequency artifacts in neighboring pixels, demonstrates better performance than UnivFD on the TwinSynths-GAN subset.
On the other hand, the generated images in TwinSynths-DM contain low-level discriminative features introduced by the DDIM decoder, which incorporates additional fully connected layers and post-processing steps following the upsampling blocks.
We can see that NPR exhibits lower performance, whereas UnivFD demonstrates higher performance.
Nevertheless, SFLD demonstrates superior and robust performance on both benchmarks, indicating its ability to capture both low-level feature artifacts and high-level feature artifacts.
Notably, no existing detector has ever exhibited such a high level of performance on both benchmarks.

\subsection{Low-level Vision and Perceptual Benchmark}
\label{sec:experiments_lowlevel}

\begin{table}[t]
    \centering
    \resizebox{0.8\linewidth}{!}{
        \renewcommand{\g}[1]{\gradientcell{#1}{50}{100}{white}{gray}{70}}

\begin{tabular}{lcccc}
\toprule
Tasks & SITD & SAN & CRN & IMLE \\
\midrule
UnivFD\cite{ojha2023towards} & \g{65.9} & \g{81.2} & \g{96.4} & \g{98.4} \\
NPR\cite{tan2024rethinking} & \g{55.2} & \g{60.0} & \g{50.0} & \g{50.0} \\
SFLD & \g{71.9} & \g{90.5} & \g{95.8} & \g{98.7} \\
\bottomrule
\end{tabular}

    }
    \caption{Low-level vision and perceptual benchmarks. Values indicate AP scores.}
    \label{tab:low-level-vision-tasks}
\end{table}

\cref{tab:low-level-vision-tasks} shows the detection performance on different benchmarks from ForenSynths\cite{wang2020cnn}.
Low-level vision models, including SITD and SAN, preserve high-level features of real images.
Perceptual models (CRN and IMLE) color semantically segmented images to match real images, preserving semantic information.
Notably, while NPR was able to detect some super-resolution images from SAN, it failed to perform well in other image-to-image translation tasks. This indicates that detectors specialized in identifying low-level feature artifacts from ProGAN struggle to generalize to images generated from different vision tasks. 
Conversely, a detector that focuses on high-level feature artifacts demonstrates strong performance on these benchmarks.
SFLD integrates semantic and structural information from different patch sizes to show superior performance on low-level vision and perceptual benchmarks.

\section{Discussion}
\subsection{Detailed Comparison with UnivFD}

\begin{table}[t]
    \centering
    \resizebox{0.8\linewidth}{!}{
        \renewcommand{\g}[1]{\gradientcell{#1}{50}{100}{white}{gray}{70}}

\begin{tabular}{lcccc}
    \toprule
    \multirow{2}*{Tasks} & \multicolumn{2}{c}{UnivFD\cite{ojha2023towards}} & \multicolumn{2}{c}{SFLD (ours)} \\
    \cmidrule(lr){2-3}\cmidrule(lr){4-5}
    & Real & Fake & Real & Fake \\
    \midrule
    ProGAN & \g{99.9} & \g{100} & \g{100} & \g{100} \\
    StyleGAN & \g{99.4} & \g{69.4} & \g{99.4} & \g{93.9} \\
    StyleGAN2 & \g{99.8} & \g{51.5} & \g{100} & \g{83.8} \\
    BigGAN & \g{98.1} & \g{92.2} & \g{93.2} & \g{99.8} \\
    CycleGAN & \g{98.9} & \g{98.4} & \g{98.3} & \g{100} \\
    StarGAN & \g{93.6} & \g{98.1} & \g{98.9} & \g{99.9} \\
    GauGAN & \g{99.3} & \g{100} & \g{92.0} & \g{100} \\
    Deepfake & \g{94.8} & \g{40.6} & \g{85.2} & \g{83.2} \\
    DALLE & \g{99.1} & \g{75.8} & \g{96.2} & \g{94.1} \\
    ADM & \g{97.2} & \g{42.8} & \g{95.3} & \g{76.6} \\
    Glide\_100\_10 & \g{99.1} & \g{57.0} & \g{96.2} & \g{85.0} \\
    Glide\_100\_27 & \g{99.1} & \g{58.2} & \g{96.2} & \g{85.2} \\
    Glide\_50\_27 & \g{99.1} & \g{59.3} & \g{96.2} & \g{88.7} \\
    LDM\_100 & \g{99.1} & \g{91.2} & \g{96.2} & \g{94.9} \\
    LDM\_200 & \g{99.1} & \g{90.0} & \g{96.2} & \g{95.2} \\
    LDM\_200\_cfg & \g{99.1} & \g{49.2} & \g{96.2} & \g{69.6} \\
    \midrule
    Avg. & \g{98.4} & \g{73.4} & \g{96.0} & \g{90.6} \\
    \bottomrule
\end{tabular}

    }
    \caption{Classification accuracy on real and fake sets on ForenSynths \cite{wang2020cnn} and diffusion sets in Ojha \etal \cite{ojha2023towards}.}
    \label{tab:real-fake-accuracy}
\end{table}

This section presents a comprehensive comparison of SFLD against UnivFD.
\cref{tab:real-fake-accuracy} shows the classification accuracy of the prediction results of real and fake images on each generator in the conventional benchmark.
It is evident that SFLD exhibits superior performance in predicting generated images.
Notably, UnivFD is unable to predict fake images in some generated subsets, whereas SFLD demonstrates its strength in both real and generated images.
This result supports that SFLD can capture both low-level feature artifacts and high-level feature artifacts, making the detector better generalize on novel generators.

\subsection{Score Ensembling}
\label{sec:score-emsembling}
\begin{figure}[t]
    \centering
    \begin{subfigure}[t]{0.45\linewidth}
        \centering
        \includegraphics[width=\textwidth]{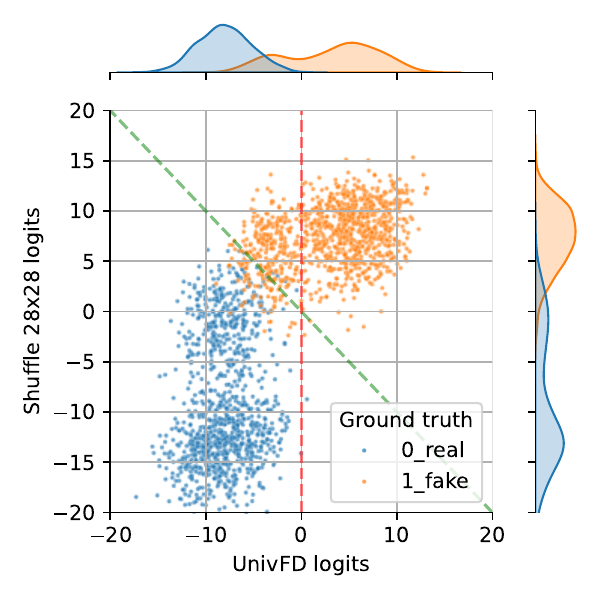}
        \subcaption{StyleGAN}
        \label{fig:scatterplot-stylegan}
    \end{subfigure}
    \begin{subfigure}[t]{0.45\linewidth}
        \centering
        \includegraphics[width=\textwidth]{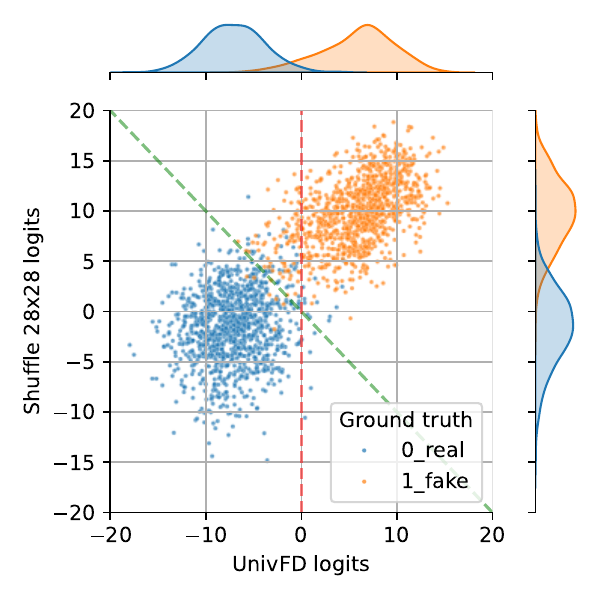}
        \subcaption{BigGAN}
    \end{subfigure}
    \begin{subfigure}[t]{0.45\linewidth}
        \centering
        \includegraphics[width=\textwidth]{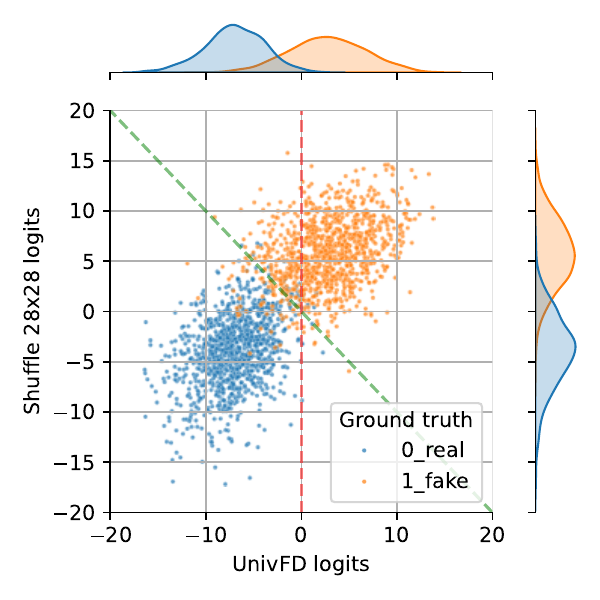}
        \subcaption{DALL-E}
    \end{subfigure}
    \begin{subfigure}[t]{0.45\linewidth}
        \centering
        \includegraphics[width=\textwidth]{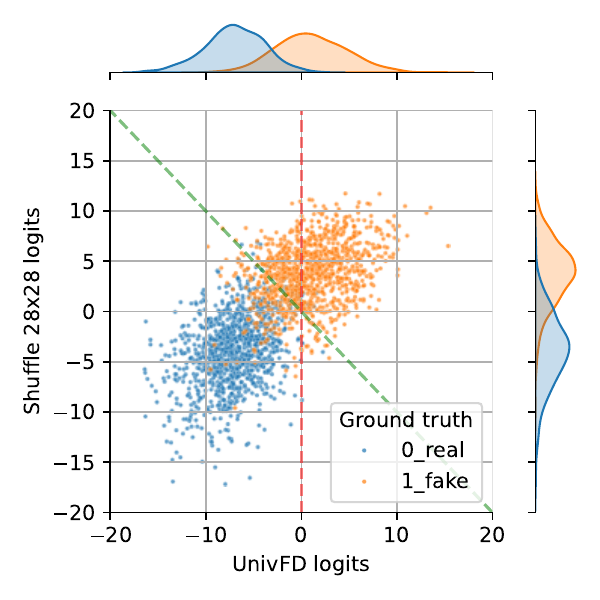}
        \subcaption{GLIDE\_50\_27}
    \end{subfigure}
     \caption{Scatter plots of per-sample scores. X-axis is the UnivFD logits, and Y-axis is the logit from PatchShuffle with patch size 28. The decision boundary of UnivFD (\textcolor{red}{red}) and SFLD (\textcolor{green}{green}) are shown. See \cref{sec:additional-results-on-scatter-plots} for extended results.}
     \label{fig:score-ensemble-viz}
\end{figure}

\textbf{Scatter plots.} 
Ensembling of the detection scores of the original image and patch-shuffled images is supported by \cref{fig:score-ensemble-viz}. In all cases, ensembling the two detectors with patch sizes 224 and 28 as an average of the two logit scores consistently improved binary separation and thus resulted in superior performance with the default threshold (as evidenced by \cref{tab:main-results-map,tab:main-results-acc}). This proves that the two detection methods work as complementary functions. 

\begin{figure}[t]
    \centering
    \begin{subfigure}{\linewidth}
        \centering
        \begin{subfigure}[t]{0.23\linewidth}
            \centering
            \includegraphics[width=\linewidth]{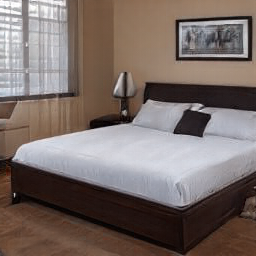}
        \end{subfigure}
        \begin{subfigure}[t]{0.23\linewidth}
            \centering
            \includegraphics[width=\linewidth]{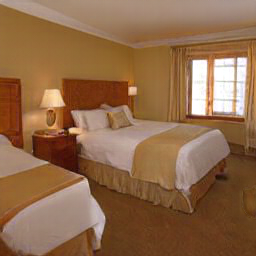}
        \end{subfigure}
        \begin{subfigure}[t]{0.23\linewidth}
            \centering
            \includegraphics[width=\linewidth]{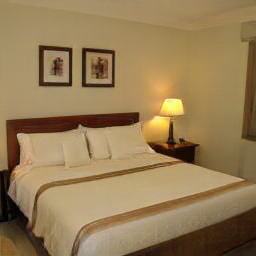}
        \end{subfigure}
        \begin{subfigure}[t]{0.23\linewidth}
            \centering
            \includegraphics[width=\linewidth]{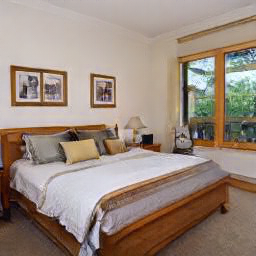}
        \end{subfigure}
        \subcaption{Fake image examples on the second quadrant of \cref{fig:scatterplot-stylegan}.}
        \label{fig:scatterplot-quadrant2}
    \end{subfigure}
    \begin{subfigure}{\linewidth}
        \centering
        \begin{subfigure}[t]{0.23\linewidth}
            \centering
            \includegraphics[width=\linewidth]{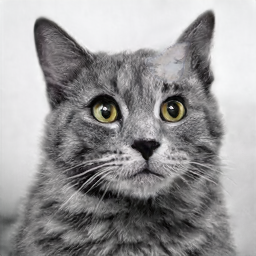}
        \end{subfigure}
        \begin{subfigure}[t]{0.23\linewidth}
            \centering
            \includegraphics[width=\linewidth]{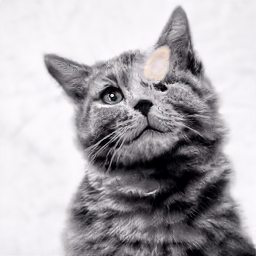}
        \end{subfigure}
        \begin{subfigure}[t]{0.23\linewidth}
            \centering
            \includegraphics[width=\linewidth]{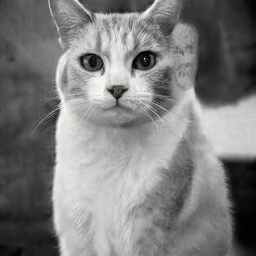}
        \end{subfigure}
        \begin{subfigure}[t]{0.23\linewidth}
            \centering
            \includegraphics[width=\linewidth]{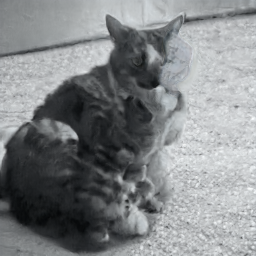}
        \end{subfigure}
        \subcaption{Fake image examples on the fourth quadrant of \cref{fig:scatterplot-stylegan}.}
        \label{fig:scatterplot-quadrant4}
    \end{subfigure}
    \caption{A closer look into the failure cases from the StyleGAN-generated test images.}
    \label{fig:scatterplot-quadrants}
\end{figure}

\textbf{A closer look into failure cases.} 
\cref{fig:scatterplot-quadrants} visualizes some exact failure cases with StyleGAN-generated images (\cref{fig:scatterplot-stylegan}).
\cref{fig:scatterplot-quadrant2} shows a case where UnivFD fails and PatchShuffle succeeds. These images seem to cause UnivFD to fail because the high-level feature is well generated (high global structure fidelity). In contrast, PatchShuffle, which focuses on local structure, succeeds in detection. 
Our method with score ensembling was able to capture these examples illustrated as the green line in \cref{fig:score-ensemble-viz}.
On the other hand, \cref{fig:scatterplot-quadrant4} shows a case where PatchShuffle fails and UnivFD succeeds. These generated images have well-generated local structures like textures but have defects in global structures such as ears, eyes, and faces. However, there are very few examples corresponding to this. This analysis indicates that using both local and global information is necessary for detecting generated images.

\subsection{Robustness Against Image Degradation}
\label{sec:discussion_robustness}

\begin{figure}[t]
    \centering
    \begin{subfigure}[t]{0.49\linewidth}
        \centering
        \includegraphics[width=1\linewidth]{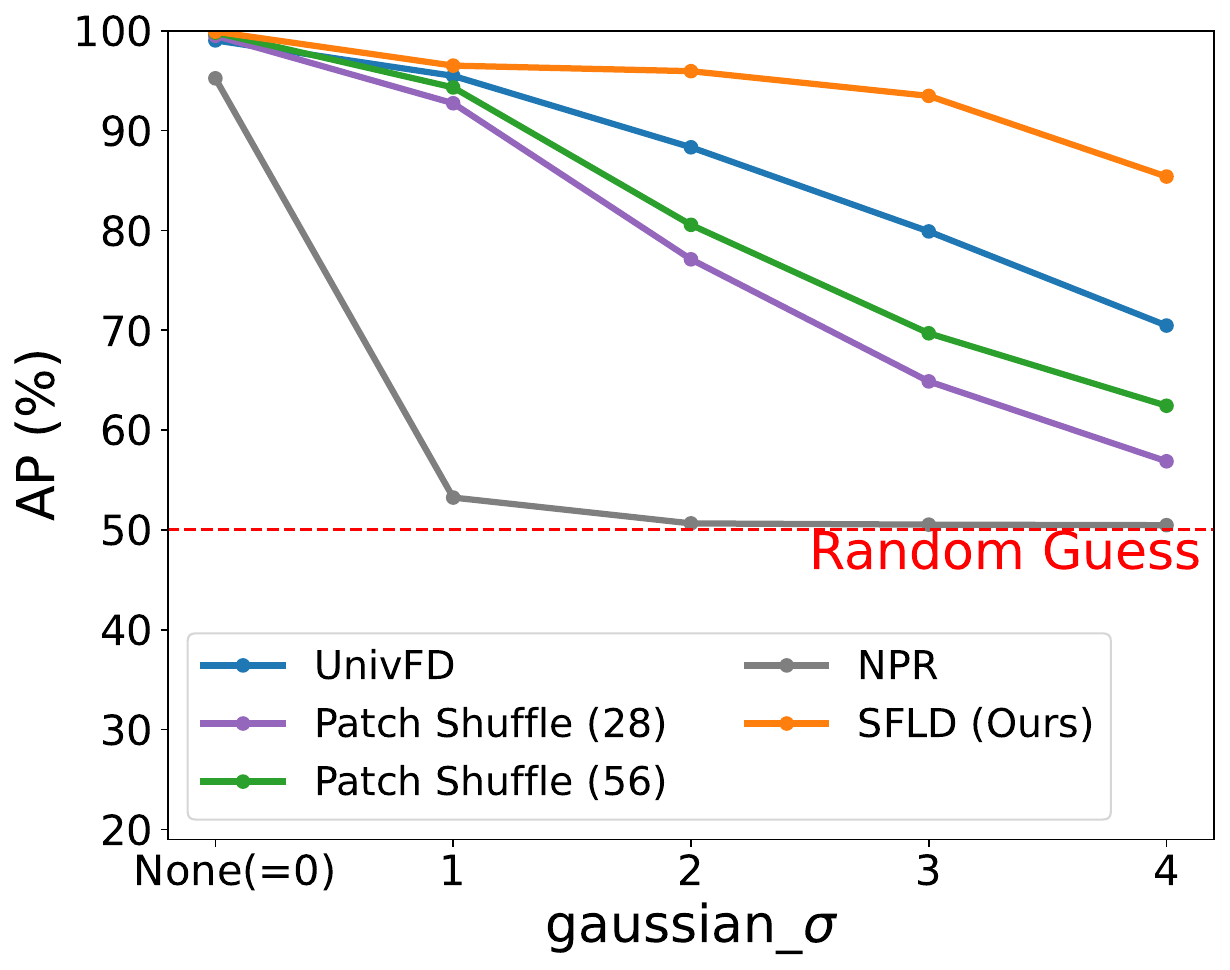}
        \subcaption{\makecell{Gaussian blur, \\GANs from \cite{wang2020cnn}.}}
        \label{fig:gan_gaussian}
    \end{subfigure}
    \begin{subfigure}[t]{0.49\linewidth}
        \centering
        \includegraphics[width=1\linewidth]{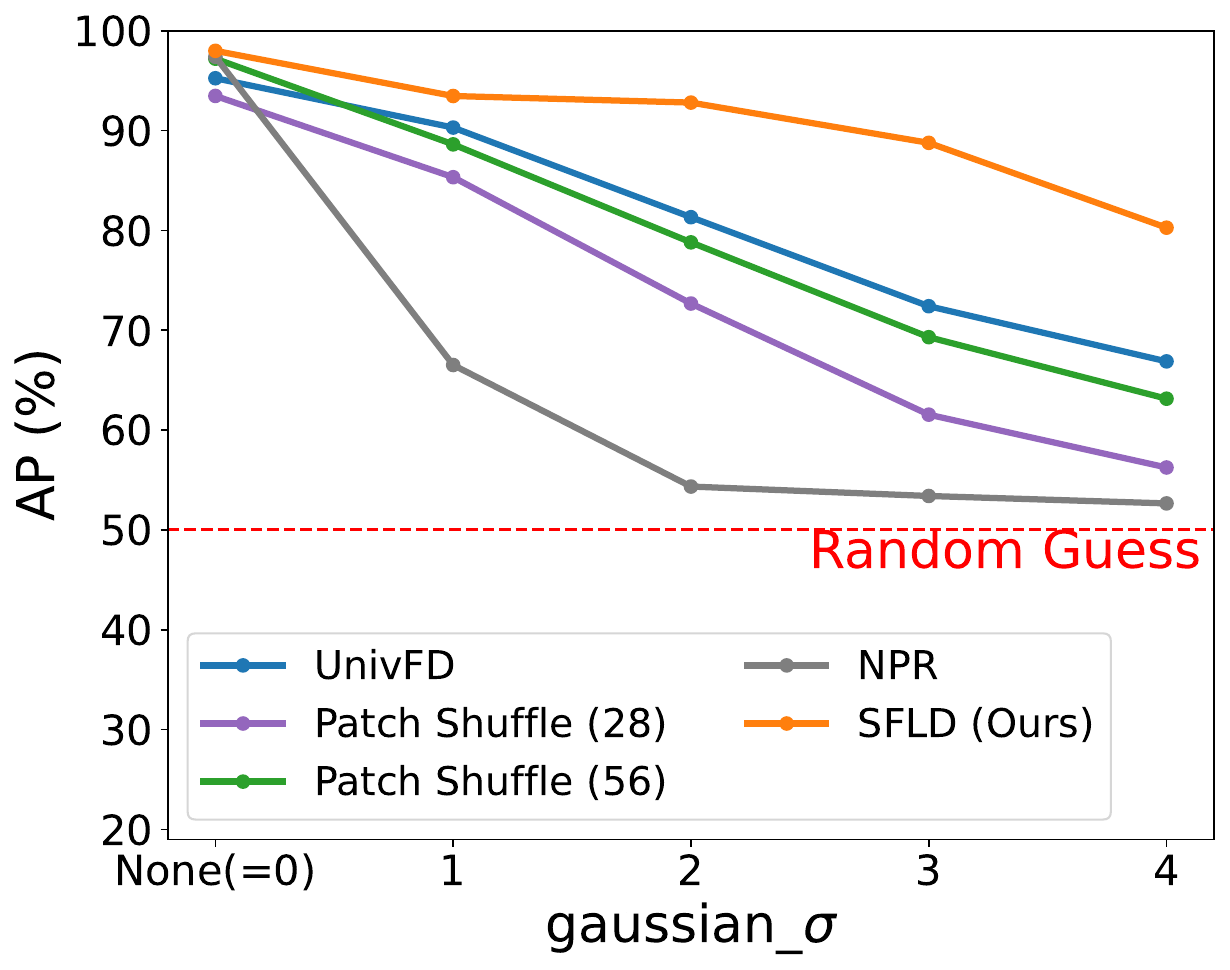}
        \subcaption{\makecell{Gaussian blur, \\DMs from \cite{ojha2023towards}.}}
        \label{fig:diff_gaussian}
    \end{subfigure}
    \begin{subfigure}[t]{0.49\linewidth}
        \centering
        \includegraphics[width=1\linewidth]{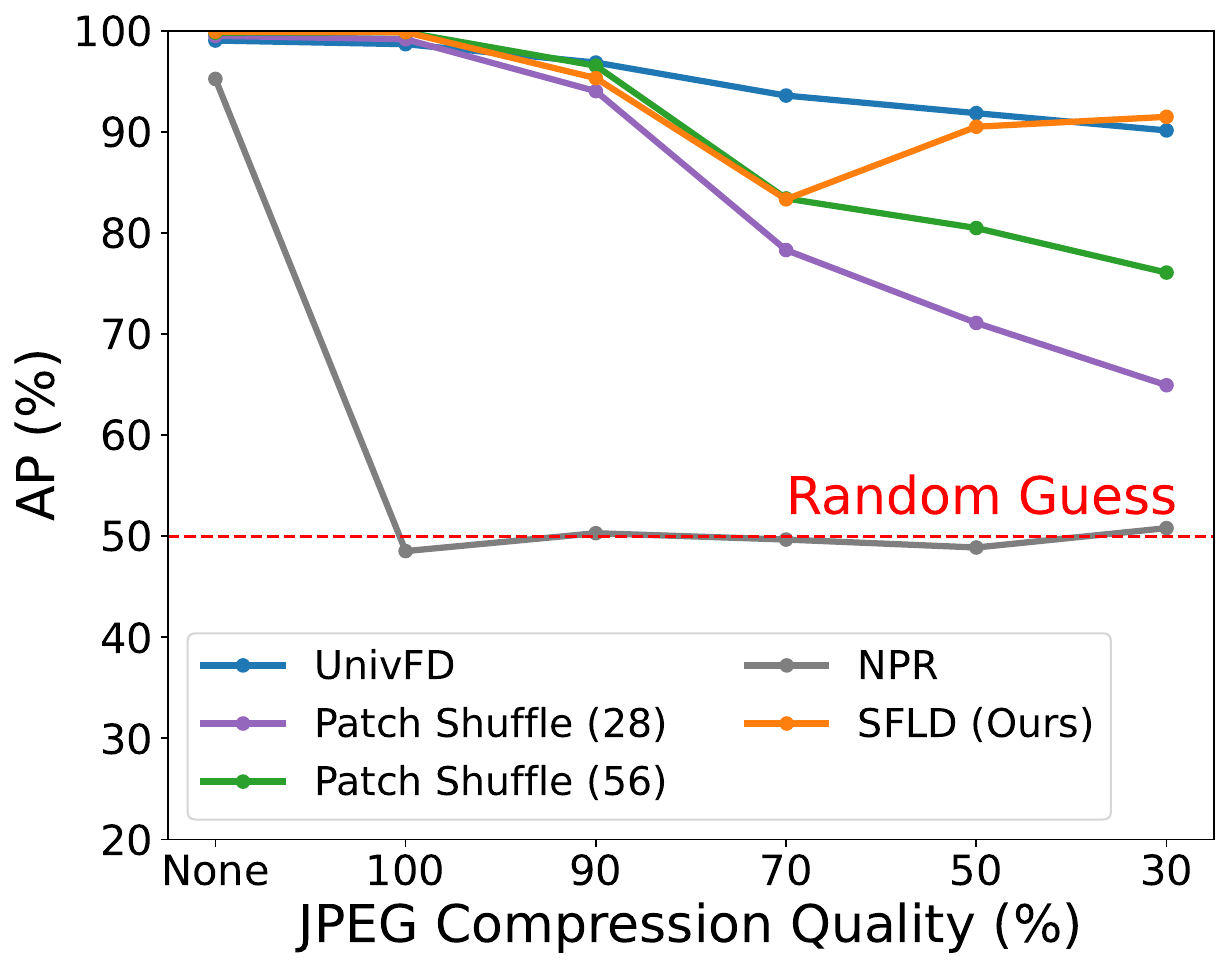}
        \subcaption{\makecell{JPEG quality, \\GANs from \cite{wang2020cnn}.}}
        \label{fig:gan_jpeg}
    \end{subfigure}
    \begin{subfigure}[t]{0.49\linewidth}
        \centering
        \includegraphics[width=1\linewidth]{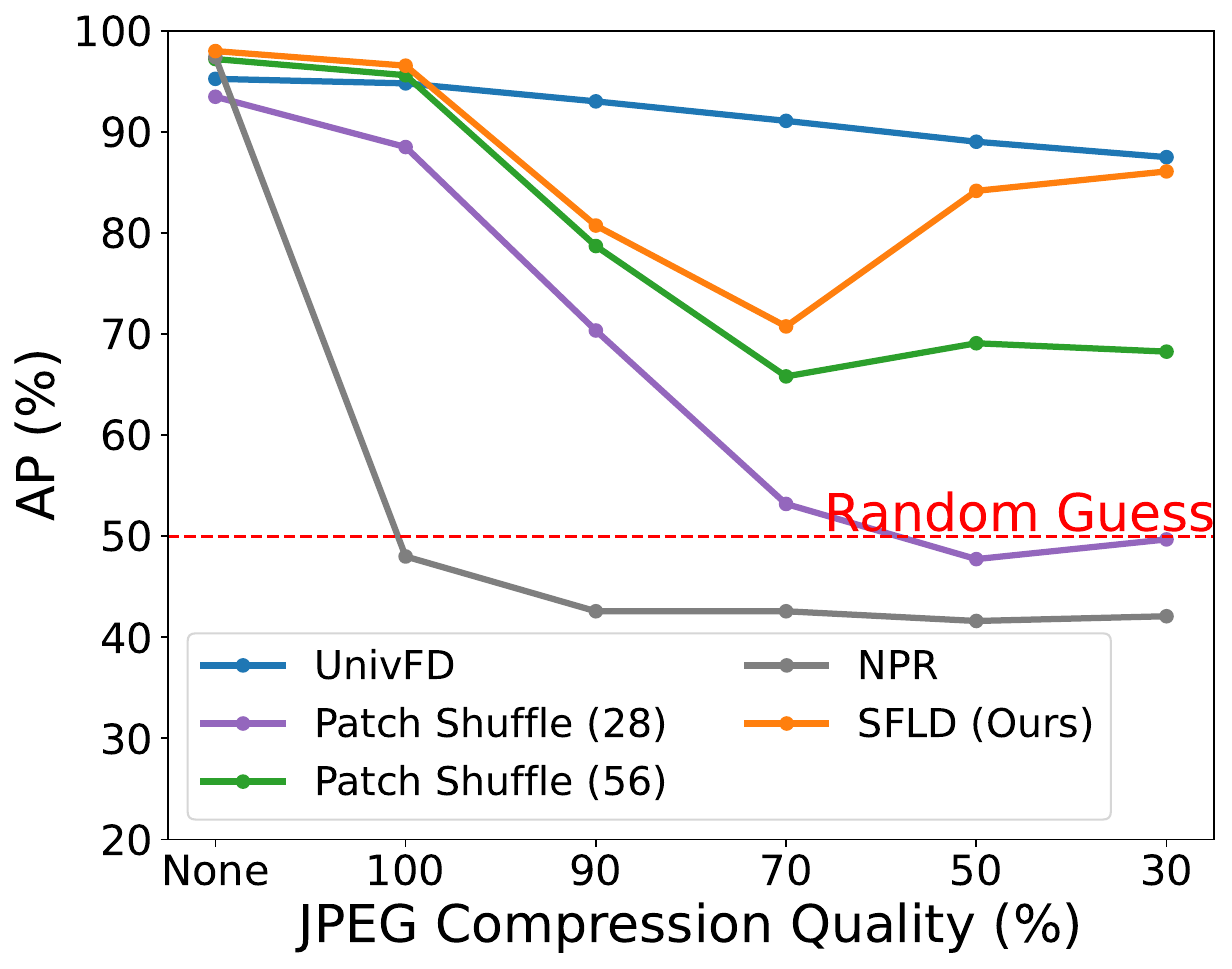}
        \subcaption{\makecell{JPEG quality, \\DMs from \cite{ojha2023towards}.}}
        \label{fig:diff_jpeg}
    \end{subfigure}
    \caption{Robustness against simulated image degradation. Methods include Gaussian blur and JPEG compression.}
    \label{fig:robustness}
\end{figure}

Applying a Gaussian blur and JPEG compression to an image is a common degradation that can naturally occur. 
\cref{fig:robustness} illustrates the impact of each attack on two subsets of generated images.
The diffusion-subset and GAN-subset are subsets of diffusion and GAN generators, respectively, drawn from the conventional benchmark.
Gaussian indicates the addition of a Gaussian blur with a standard deviation of $\sigma$.
JPEG indicates the application of JPEG compression with a specified compression quality.
Note that JPEG compression with quality 100 does not result in the same image, as JPEG compression reduces color information and rounds coefficients, thereby losing some information.

If the model is vulnerable to image degradation, we can infer that it is influenced by the features targeted by the degradation.
Specifically, Gaussian blur affects both high- and low-level features in the image, while JPEG compression primarily targets low-level features (see \cref{fig:robustness_examples}).
\cref{fig:gan_gaussian,fig:diff_gaussian} demonstrates that SFLD always shows the best performance against Gaussian blur, since it integrates both high- and low-level features through ensemble/fusion, enabling each to compensate for the information lost in the other.
\cref{fig:gan_jpeg,fig:diff_jpeg} illustrates that SFLD restores robustness against JPEG compression, supporting the fundamental principle behind our model.
Additionally, UnivFD, which focuses on capturing high-level feature artifacts is also robust against JPEG compression.
However, NPR, which focuses on capturing low-level feature artifacts, is vulnerable to both Gaussian blur and JPEG compression even at JPEG compression quality 100.

\subsection{Qualitative Analysis}
\label{sec:qualitative-analysis}

\begin{figure}[t]
    \centering
    \begin{subfigure}[b]{\linewidth}
        \centering
        \includegraphics[width=\textwidth]{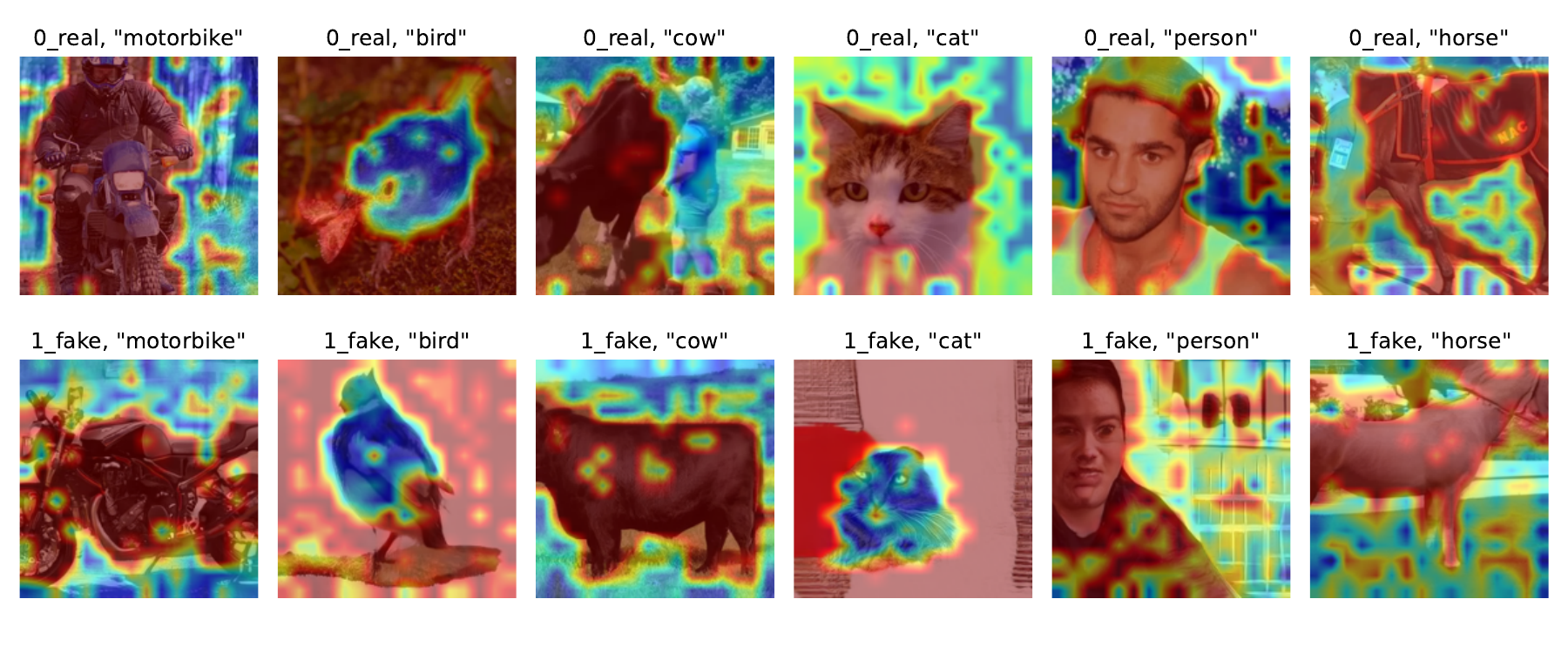}
        \caption{UnivFD\cite{ojha2023towards} examples}
    \end{subfigure}
    \begin{subfigure}[b]{\linewidth}
        \centering
        \includegraphics[width=\textwidth]{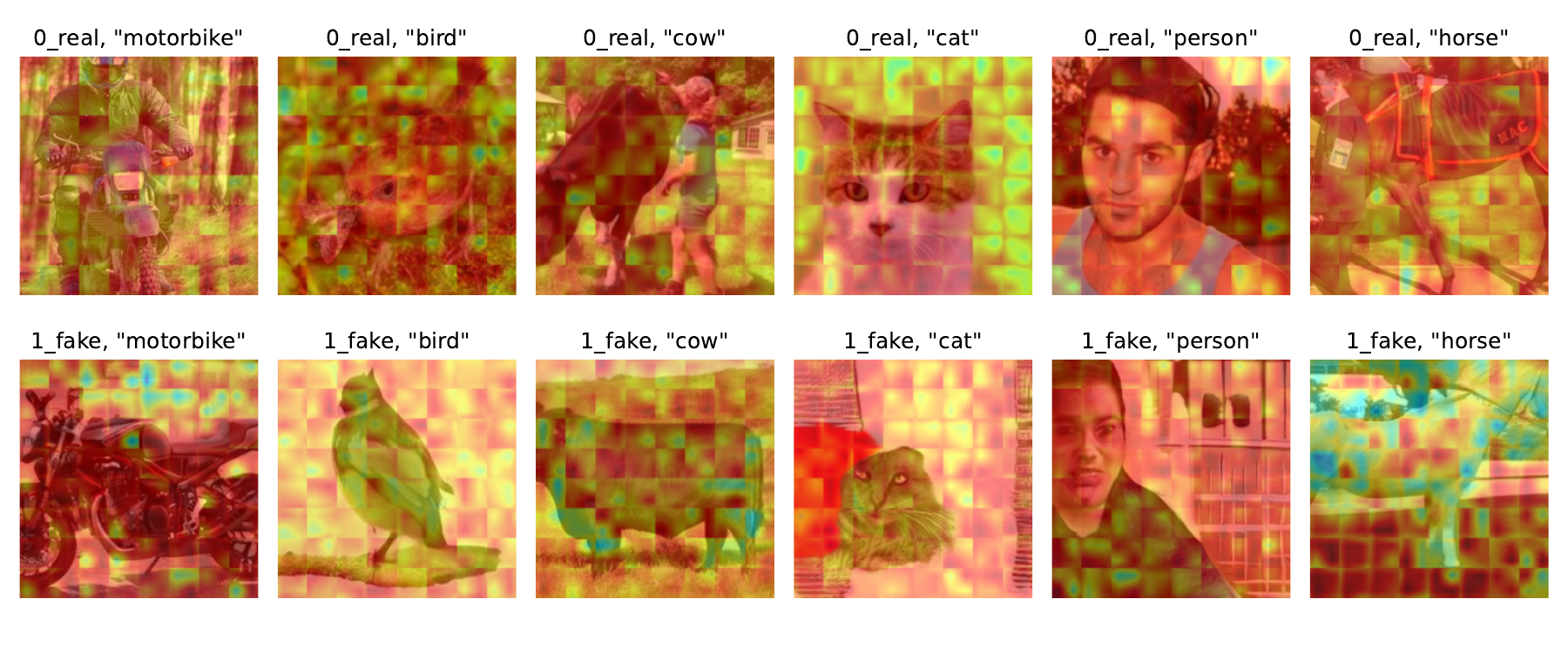}
        \caption{Patch-shuffle 28×28 examples}
        \label{fig:cam-viz-ps28x28}
    \end{subfigure}
    \caption{Class activation maps (CAM) for UnivFD\cite{ojha2023towards} and the patch-shuffled detector (ours). GradCAM\cite{selvaraju2020grad,jacobgilpytorchcam} was used to obtain the heatmaps. The ground truth real/fake labels and class labels are displayed on top of each image. Note that for \cref{fig:cam-viz-ps28x28}, the heat map is split into patches then reverse-shuffled back to the corresponding spatial location of the input image. }
    \label{fig:cam-viz}
\end{figure}

\textbf{GradCAM visualization.} 
See \cref{fig:cam-viz} for image attribution heat maps generated using GradCAM\cite{selvaraju2020grad,jacobgilpytorchcam}.
The examples are from the ProGAN test set.
In addition, the heat maps are averaged across ten predictions to reduce the randomness from the patch permutation. 
The CAM of UnivFD focuses on the class-dependent salient region, whereas the patch-shuffled detector focuses on the entire image region.

\begin{figure}[t]
  \centering
  \begin{subfigure}[t]{\linewidth}
    \centering
    \includegraphics[width=\linewidth]{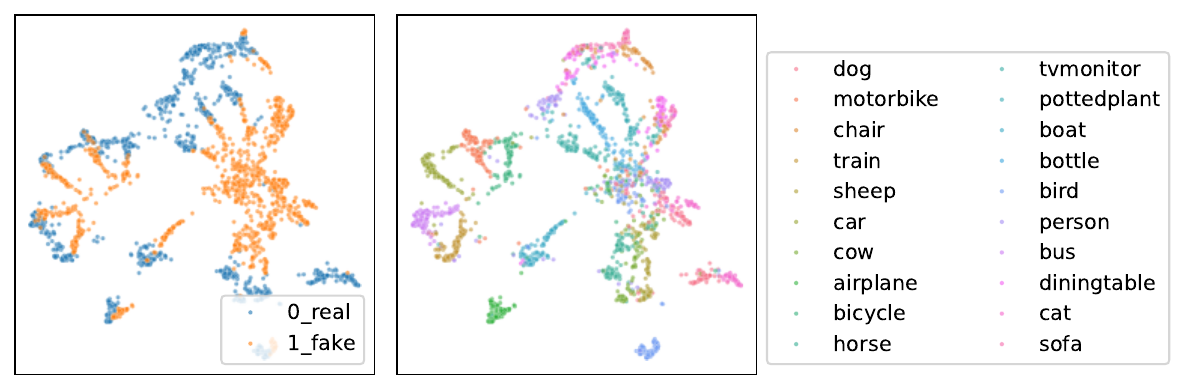}
    \subcaption{Embeddings from UnivFD\cite{ojha2023towards,CLIP}}
  \end{subfigure}
  \begin{subfigure}[t]{\linewidth}
    \centering
    \includegraphics[width=\linewidth]{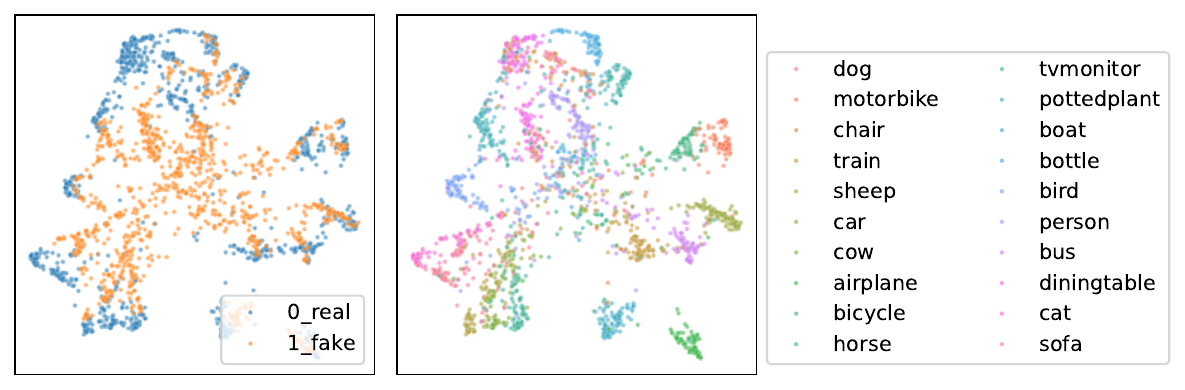}
    \subcaption{PatchShuffle(28) embeddings averaged across $N_{\text{views}}=10$ shuffles.}
  \end{subfigure}
  \caption{UMAP visualization\cite{SMG2020} of feature embeddings. Left and right plots show the same projected embeddings colored by real/fake labels (left) and object category labels (right). Our method destroys the class information from the embeddings, thereby improving the generalization by reducing the content bias.}
  \label{fig:embedding-visualization}
\end{figure}

\textbf{Feature visualization.}
Because taking an average of the logits generated via a linear layer is equivalent to taking an average of the feature embeddings, we can understand the SFLD embeddings by taking the average of the embeddings over multiple shuffles. \cref{fig:embedding-visualization} visualizes the feature embeddings by projecting onto a 2D plane using UMAP\cite{SMG2020}. We used the ProGAN test set to extract the embeddings. 

Because UnivFD learns the features directly from the CLIP visual encoder, the embeddings form class-dependent clusters. This creates class-dependent decision boundary, which may introduce unintended content bias to the real-fake detector. In contrast, because PatchShuffle destroys class-related information from the image, the corresponding embeddings show more dispersion within each class.

\subsection{Effect of PatchShuffle Hyperparameters}

\textbf{Improving feature extraction with PatchShuffle.}
We suggest additional details to get better CLIP features from the shuffled images.
To improve stability against the randomness introduced by PatchShuffle, we use the averaged logits of $N_{views}=10$ randomly shuffled patch combinations for each input image during testing.

Moreover, in our problem setup, training images are fixed at 256×256 size, while test images can vary in size.
Resizing test images is avoided, as image degradation due to resizing (e.g., JPEG compression or blur) has been shown to impact the detection of AI-generated images negatively \cite{wang2020cnn}.
Instead, recent detectors\cite{tan2024rethinking, ojha2023towards} prefer cropping over resizing.
Our backbone model without PatchShuffle also extracts CLIP features from 224×224 center-cropped images without resizing. 
However, we can extract information not only from the center of the image but from the entire image by taking advantage of the proposed PatchShuffle, which allows non-consecutive patchwise combinations.
We divide the entire test image into non-overlapping patches of the given patch size and combine these patches into 224×224 images. This approach enables the detector to analyze information from the entire image, rather than being constrained to a single central region. 
See \cref{sec:Appendix-fullimagesampling} for more details.

\begin{figure}[t]
    \centering
    \begin{subfigure}[t]{0.45\linewidth}
        \centering
        \includegraphics[width=\linewidth]{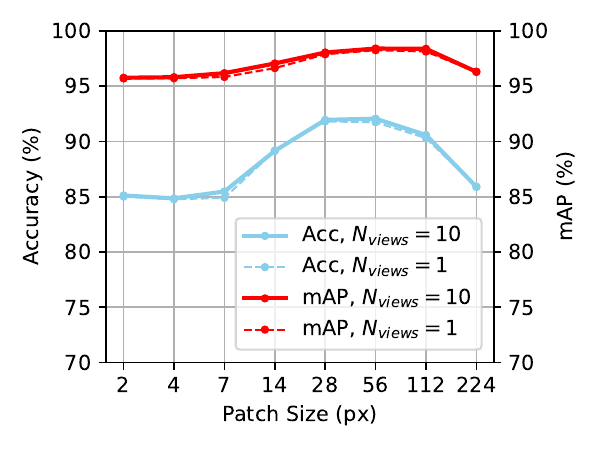}
        \subcaption{Sweep over patch size}
        \label{fig:patch_size}
    \end{subfigure}
    \begin{subfigure}[t]{0.45\linewidth}
        \centering
        \includegraphics[width=\linewidth]{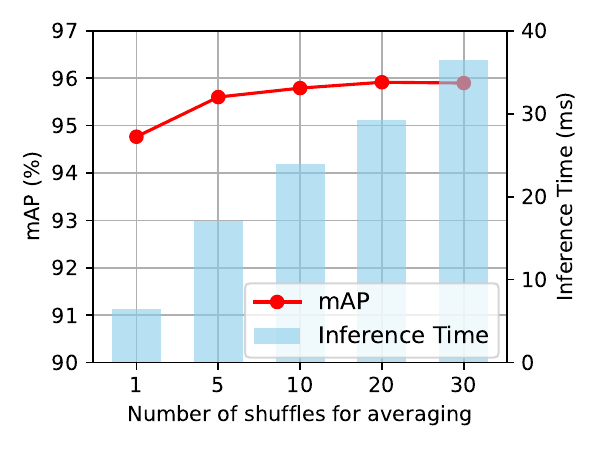}
        \subcaption{Sweep over $N_{\text{views}}$}
        \label{fig:view}
    \end{subfigure}
    \caption{Best patch sizes were found at 28×28 and 56×56. $N_{\text{views}}=10$ showed the best balance between performance and inference cost.}
\end{figure}

\label{subsec:patchsetting}
\textbf{Patch size.}
The optimal patch size should be sufficiently small to disrupt the underlying image structure while preserving some high-level feature artifacts. The results for the performance difference according to patch sizes on a conventional benchmark are presented in \cref{fig:patch_size}. Each patch size model in x-axis refers to the ensemble between the corresponding PatchShuffle model and UnivFD(patch size 224). It can be observed that an too small patch size and an excessively large patch size do not assist the model in capturing useful high-level and low-level feature artifacts. Therefore, the majority of experiments in this paper utilized patch sizes 28x28 and 56x56 according to this result.

\textbf{Number of shuffled views.} 
To ensure the stability of the random patch shuffle, SFLD generates multiple versions of shuffled image from a single test image and employs the average of them as the score. As illustrated in \cref{fig:view}, mAP enhances with higher $N_{\text{views}}$. However, due to the tradeoff with inference time, we chose $N_{\text{views}}=10$, and all results presented in this paper were obtained with this setting. The results in \cref{fig:view} are from the PatchShuffle model with a patch size of 28, without an ensemble with UnivFD. The inference time was measured using RTX 4090 GPU.

\section{Conclusions}
\label{sec:conclusion}
In this paper, we introduced SFLD, a novel method for detecting AI-generated images that effectively combines global semantic structures and textural structures to improve detection performance. 
By leveraging random patch shuffling and an ensemble of classifiers trained on patches of varying sizes, our approach effectively addresses the shortcomings of existing methods, such as their content bias and susceptibility to image perturbations.
Also, We proposed a new quality-ensuring benchmark, TwinSynths. It is the first to consider a scenario of infinitely real-like fake images, providing a valuable resource for future research in this area. 
We demonstrated that SFLD outperforms SOTA methods in generalization to various generators, even in challenging scenarios simulated with TwinSynths.

\noindent
\textbf{Acknowledgements} This work was supported by Institute of Information \& communications Technology Planning \& Evaluation (IITP) grant funded by the Korea government(MSIT) (No.RS-2021-II212068, Artificial Intelligence Innovation Hub).

{\small
\bibliographystyle{ieee_fullname}
\bibliography{egbib}
\nocite{cazenavette2024fakeinversion}
}

\clearpage
\appendix
\onecolumn
\section{Additional results on scatter plots} 
\label{sec:additional-results-on-scatter-plots} 
\begin{figure*}[!ph]
    \centering
    \begin{subfigure}[b]{0.20\linewidth}
        \centering
        \includegraphics[width=\textwidth]{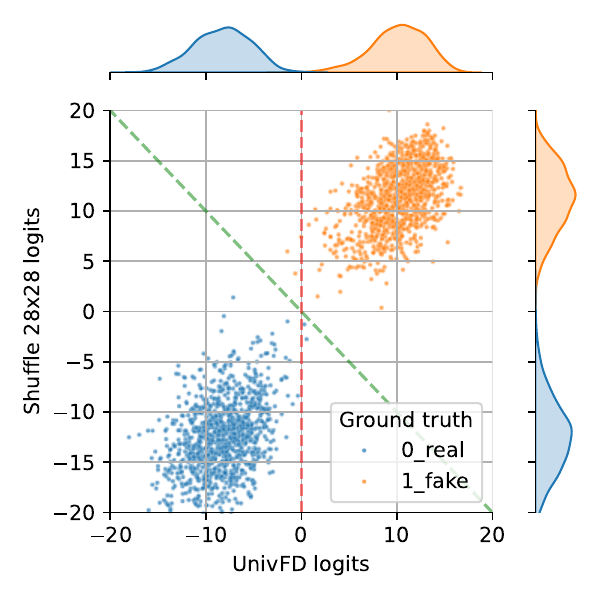}
        \subcaption{ProGAN}
    \end{subfigure}
    \begin{subfigure}[b]{0.20\linewidth}
        \centering
        \includegraphics[width=\textwidth]{figs/scatterplots/scatterplot-stylegan-shuffle28.pdf}
        \subcaption{StyleGAN}
    \end{subfigure}
    \begin{subfigure}[b]{0.20\linewidth}
        \centering
        \includegraphics[width=\textwidth]{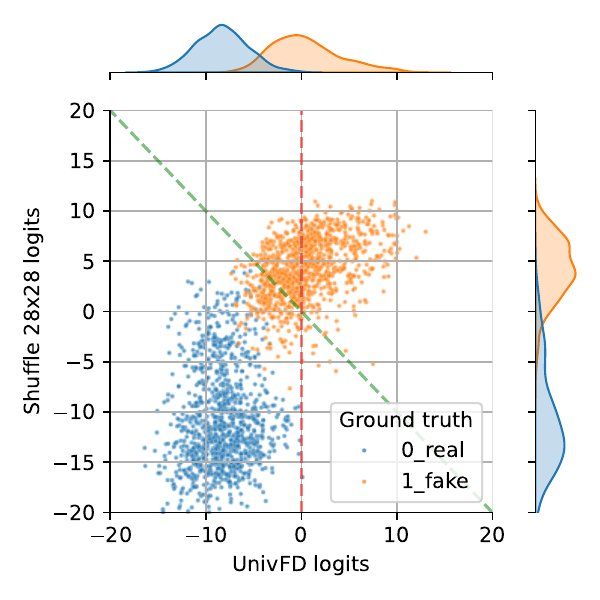}
        \subcaption{StyleGAN2}
    \end{subfigure}
    \begin{subfigure}[b]{0.20\linewidth}
        \centering
        \includegraphics[width=\textwidth]{figs/scatterplots/scatterplot-biggan-shuffle28.pdf}
        \subcaption{BigGAN}
    \end{subfigure}
    \begin{subfigure}[b]{0.20\linewidth}
        \centering
        \includegraphics[width=\textwidth]{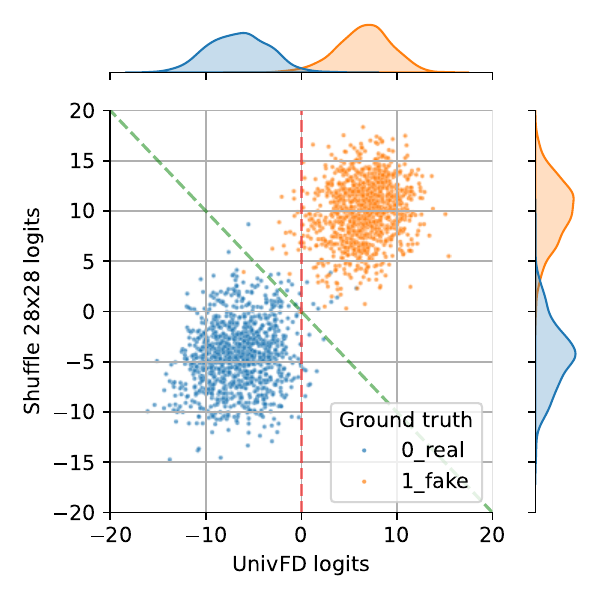}
        \subcaption{CycleGAN}
    \end{subfigure}
    \begin{subfigure}[b]{0.20\linewidth}
        \centering
        \includegraphics[width=\textwidth]{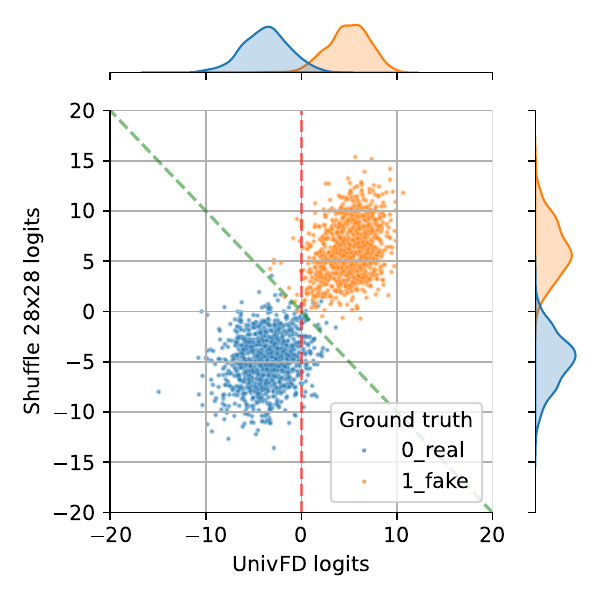}
        \subcaption{StarGAN}
    \end{subfigure}
    \begin{subfigure}[b]{0.20\linewidth}
        \centering
        \includegraphics[width=\textwidth]{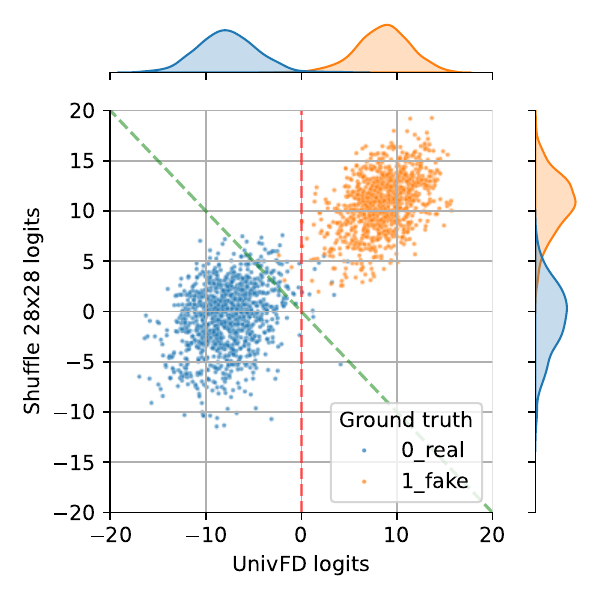}
        \subcaption{GauGAN}
    \end{subfigure}
    \begin{subfigure}[b]{0.20\linewidth}
        \centering
        \includegraphics[width=\textwidth]{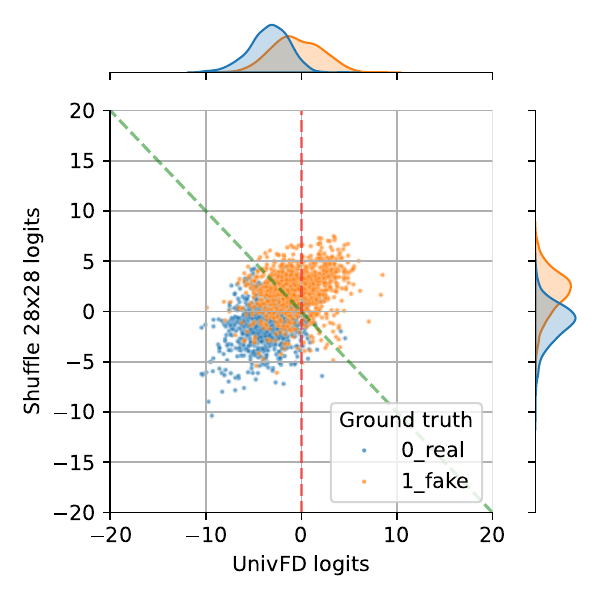}
        \subcaption{DeepFake}
    \end{subfigure}
    \begin{subfigure}[b]{0.20\linewidth}
        \centering
        \includegraphics[width=\textwidth]{figs/scatterplots/scatterplot-dalle-shuffle28.pdf}
        \subcaption{DALL-E}
    \end{subfigure}
    \begin{subfigure}[b]{0.20\linewidth}
        \centering
        \includegraphics[width=\textwidth]{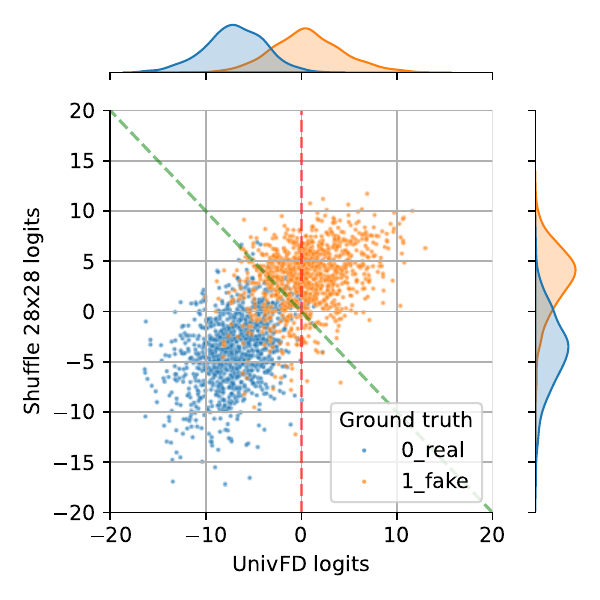}
        \subcaption{GLIDE\_100\_10}
    \end{subfigure}
    \begin{subfigure}[b]{0.20\linewidth}
        \centering
        \includegraphics[width=\textwidth]{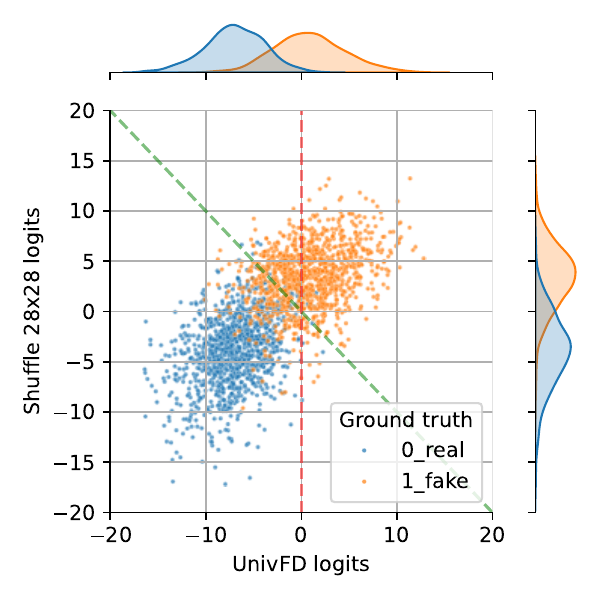}
        \subcaption{GLIDE\_100\_27}
    \end{subfigure}
    \begin{subfigure}[b]{0.20\linewidth}
        \centering
        \includegraphics[width=\textwidth]{figs/scatterplots/scatterplot-glide_50_27-shuffle28.pdf}
        \subcaption{GLIDE\_50\_27}
    \end{subfigure}
    \begin{subfigure}[b]{0.20\linewidth}
        \centering
        \includegraphics[width=\textwidth]{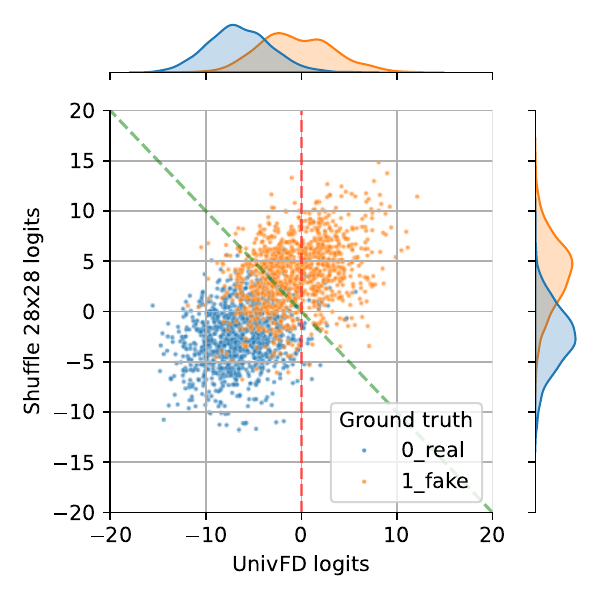}
        \subcaption{ADM}
    \end{subfigure}
    \begin{subfigure}[b]{0.20\linewidth}
        \centering
        \includegraphics[width=\textwidth]{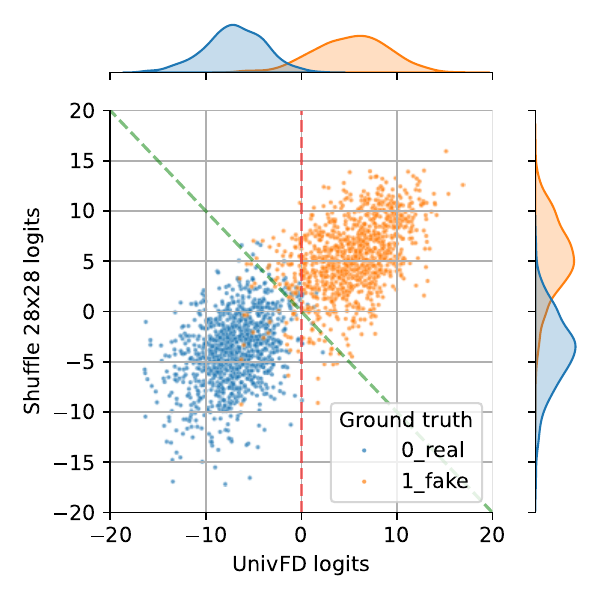}
        \subcaption{LDM\_100}
    \end{subfigure}
    \begin{subfigure}[b]{0.20\linewidth}
        \centering
        \includegraphics[width=\textwidth]{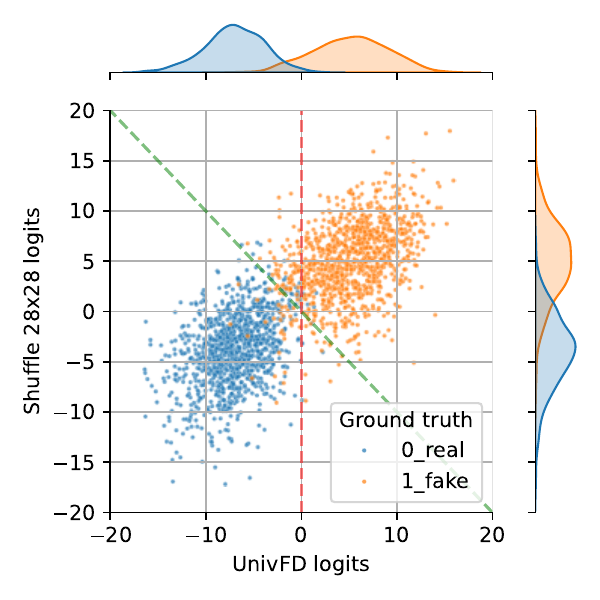}
        \subcaption{LDM\_200}
    \end{subfigure}
    \begin{subfigure}[b]{0.20\linewidth}
        \centering
        \includegraphics[width=\textwidth]{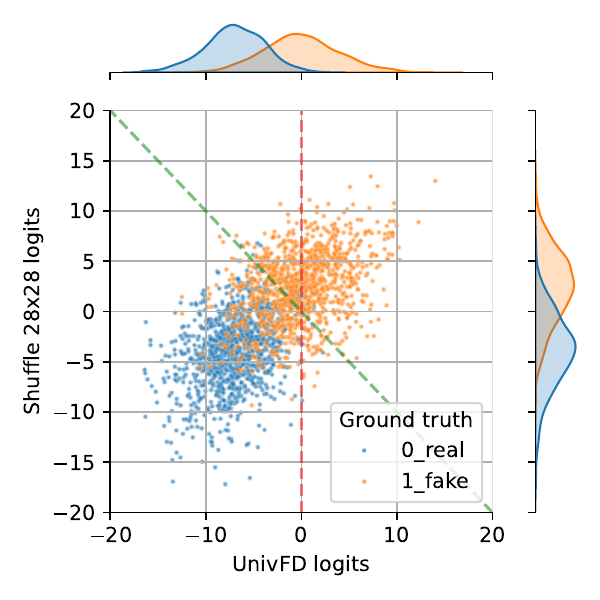}
        \subcaption{LDM\_200\_cfg}
    \end{subfigure}
    \begin{subfigure}[b]{0.20\linewidth}
        \centering
        \includegraphics[width=\textwidth]{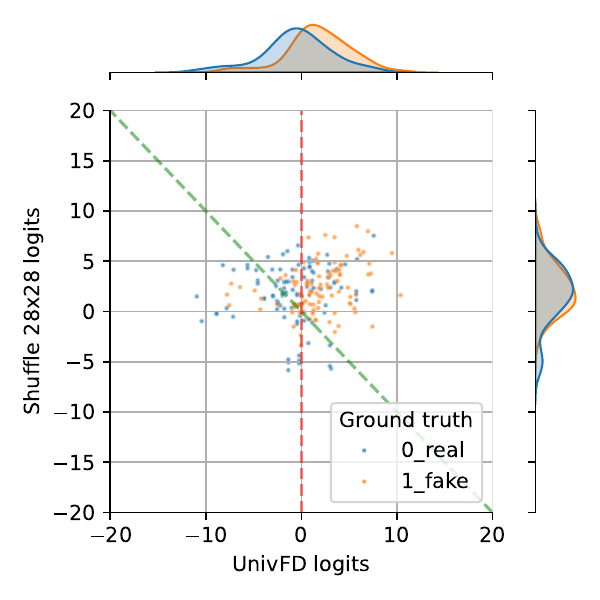}
        \subcaption{SITD}
    \end{subfigure}
    \begin{subfigure}[b]{0.20\linewidth}
        \centering
        \includegraphics[width=\textwidth]{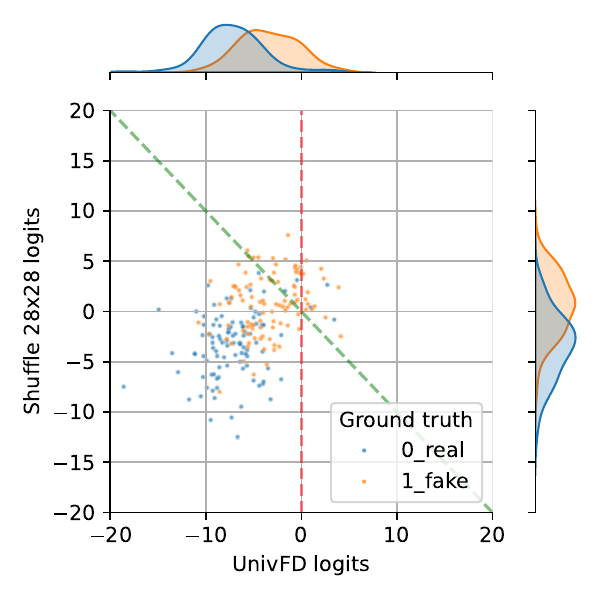}
        \subcaption{SAN}
    \end{subfigure}
    \begin{subfigure}[b]{0.20\linewidth}
        \centering
        \includegraphics[width=\textwidth]{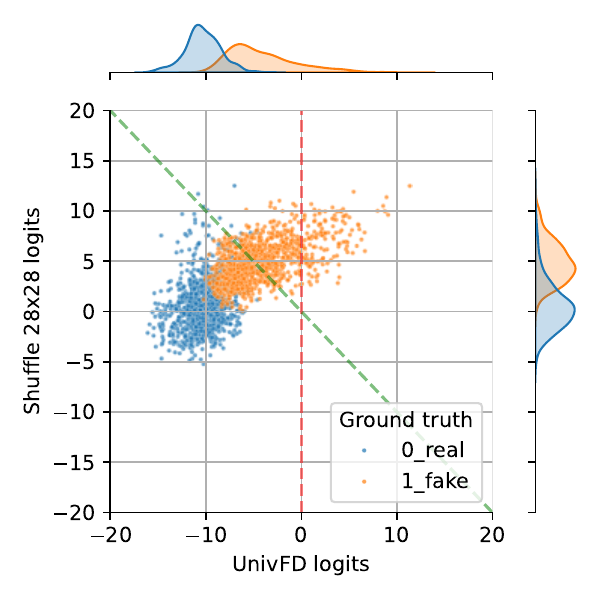}
        \subcaption{CRN}
    \end{subfigure}
    \begin{subfigure}[b]{0.20\linewidth}
        \centering
        \includegraphics[width=\textwidth]{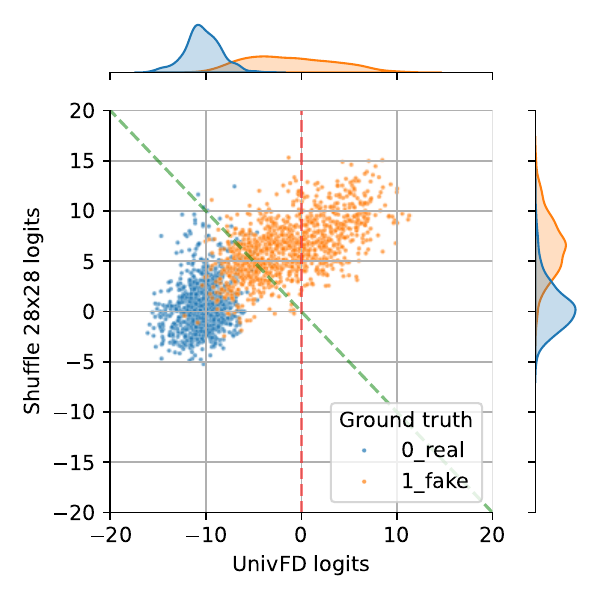}
        \subcaption{IMLE}
    \end{subfigure}
     \caption{Scatter plots of per-sample scores. X-axis is UnivFD logits, and Y-axis is the logit from PatchShuffle with patch size 28. The decision boundary of UnivFD (\textcolor{red}{red}) and SFLD (\textcolor{green}{green}) are shown.}
     \vspace{-1em}
     \label{fig:additional-score-ensemble-viz}
     \vspace{-1em}
\end{figure*}

Additional results to \cref{sec:score-emsembling} are presented in \cref{fig:additional-score-ensemble-viz}.

\twocolumn

\section{Datasets}
\label{supp:datasets}

\subsection{Train dataset}
To establish a baseline for comparison, we adopt the most common setting for training the detection model, namely the train set from ForenSynths\cite{wang2020cnn}. 
The train set consists of real images and ProGAN\cite{karras2018progressive}-generated images. It involves 20 different object class categories, each containing 18K real images from the different LSUN\cite{yu2015lsun} datasets and 18K synthetic images generated by ProGAN.

\subsection{Test dataset}
We evaluate the performance of SFLD on (1) conventional benchmarks, (2) TwinSynths which we proposed, (3) low-level vision and perceptual loss benchmarks. In this section, we provide a detailed description of the configurations for the conventional benchmarks and low-level vision and perceptual loss benchmarks.

\textbf{Conventional benchmark} 
This is from ForenSynths\cite{wang2020cnn} and Ojha \etal\cite{ojha2023towards}, including 16 different subsets of generated images, synthesized by seven GAN-based generative models, eight diffusion-based generative models and one deepfake model. 
The subset of GAN-based fake images are from ForenSynths\cite{wang2020cnn}, including ProGAN\cite{karras2018progressive}, StyleGAN\cite{karras2019style}, StyleGAN2\cite{Karras2019stylegan2}, BigGAN\cite{brock2018biggan}, CycleGAN\cite{CycleGAN}, StarGAN\cite{choi2018stargan}, and GauGAN\cite{GauGAN}.
The subset of diffusion-based fake images are from Ojha \etal\cite{ojha2023towards}, including DALL-E\cite{dayma2021dall}, three different variants of Glide\cite{nichol2021glide}, ADM(guided-diffusion)\cite{dhariwal2021diffusion}, and three different variants of LDM\cite{rombach2022high}.
Deepfake set is from FaceForensices++\cite{Deepfake} which is included in ForenSynths\cite{wang2020cnn}.
The real images corresponding to the fake images described above were directly taken from the same datasets. Those are sampled from LSUN\cite{yu2015lsun}, ImageNet\cite{russakovsky2015imagenet}, CycleGAN\cite{CycleGAN}, CelebA\cite{CelebA}, COCO\cite{coco}, and FaceForensics++\cite{Deepfake}.

\textbf{Low-level vision and perceptual loss benchmarks} 
Low-level vision benchmark consists of SITD\cite{chen2018SITD} and SAN\cite{dai2019SAN}.
These are image processing models that approximate long exposures in low light conditions from short exposures in raw camera input or process super-resolution on low-resolution images.
Perceptual benchmark consists of CRN\cite{chen2017CRN} and IMLE\cite{li2019IMLE}.
These models color the semantic segmentation map into a realistic image while directly optimizing a perceptual loss.
These benchmarks are from ForenSynths\cite{wang2020cnn}.

\section{Qualitative analysis on TwinSynths dataset}
\begin{figure}[h]
    \centering
    \begin{subfigure}[t]{\linewidth}
        \centering
        \includegraphics[width=\textwidth]{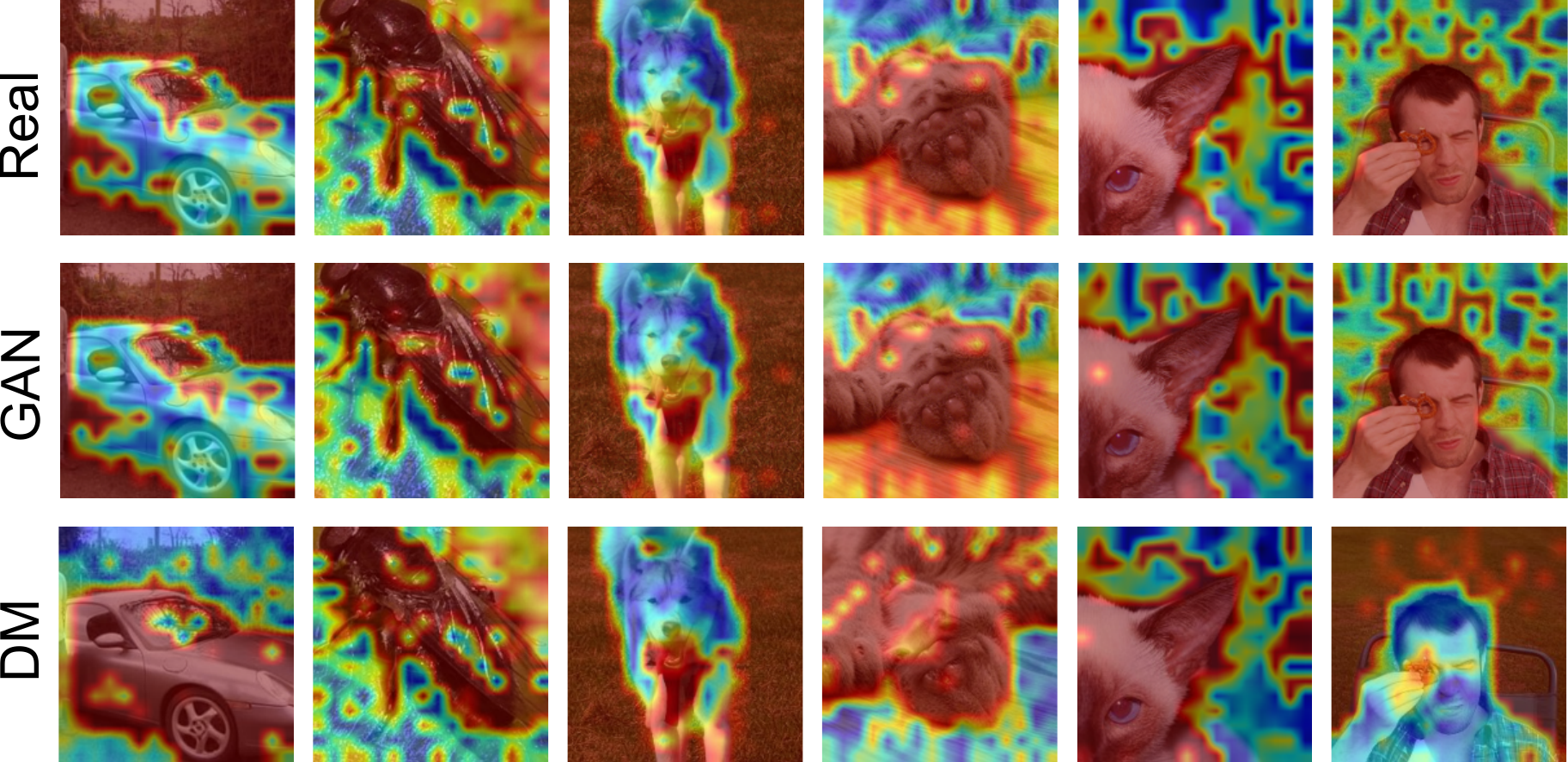}
        \caption{UnivFD\cite{ojha2023towards} examples}
    \end{subfigure}
    \begin{subfigure}[t]{\linewidth}
        \centering
        \includegraphics[width=\textwidth]{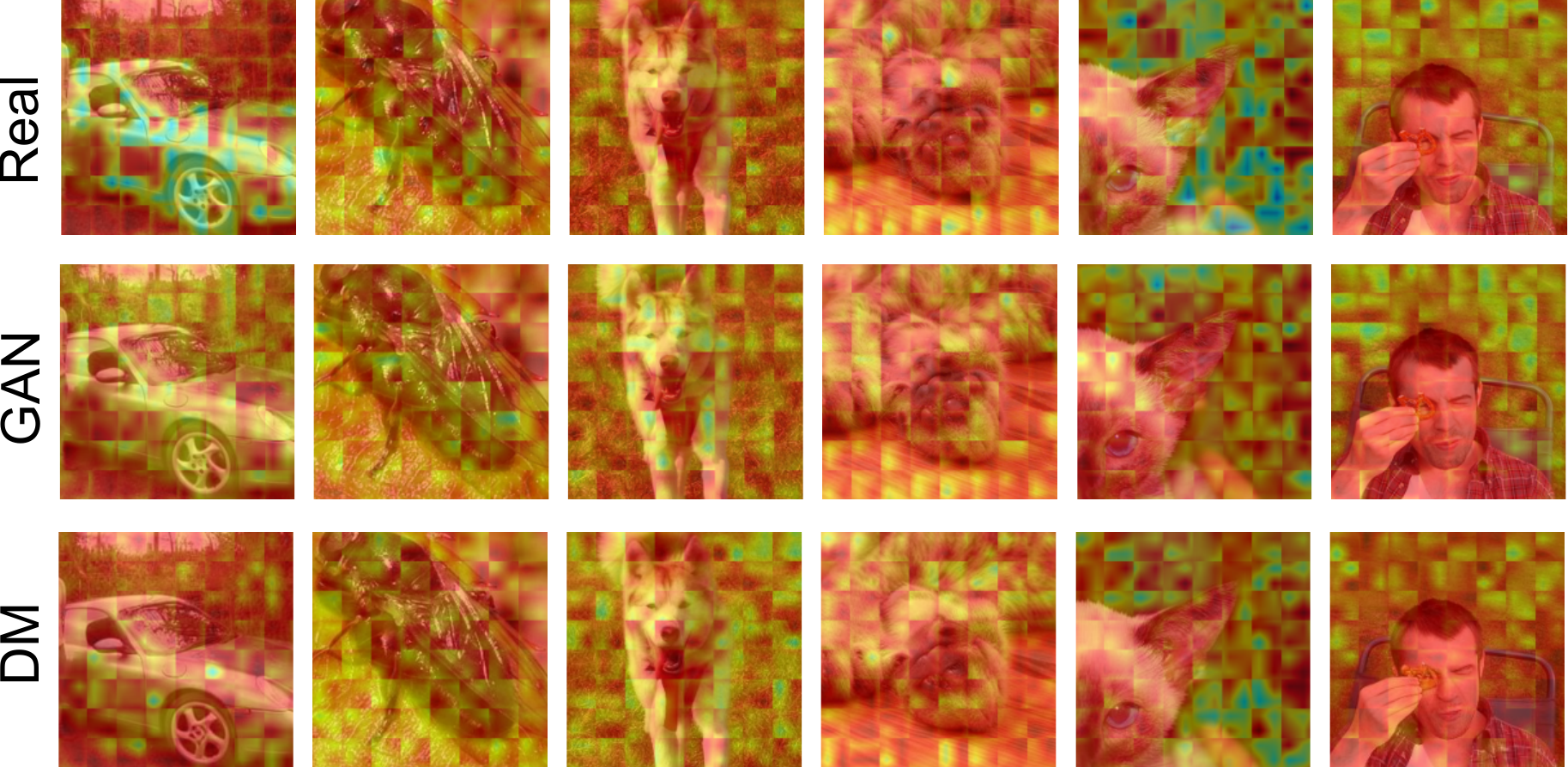}
        \caption{PatchShuffle(patch size 28) examples}
    \end{subfigure}
    \caption{Class activation maps (CAM) for UnivFD\cite{ojha2023towards} and the patch-shuffled detector (ours) in TwinSynths dataset. 
    Each row shows examples from TwinSynths-real, TwinSynths-GAN, TwinSynths-DM sets. 
    GradCAM\cite{selvaraju2020grad,jacobgilpytorchcam} was used to obtain the heatmaps. }
    \label{fig:cam-viz-twinsynth}
\end{figure}

We show the GradCAM visualization of UnivFD\cite{ojha2023towards} and Patch-shuffle 28×28 using the TwinSynths dataset in \cref{fig:cam-viz-twinsynth}.
Similar to \cref{sec:qualitative-analysis}, UnivFD is shown to focus on the class-dependent salient region, whereas our method focuses on the entire image region. 
Moreover, we observed that for TwinSynths dataset, UnivFD does respond identically to real/fake images which indicate its inability to capture subtle fake image fingerprints, whereas our method shows the response to such a difference. 

\begin{figure}[t]
    \centering
    \includegraphics[width=0.9\linewidth]{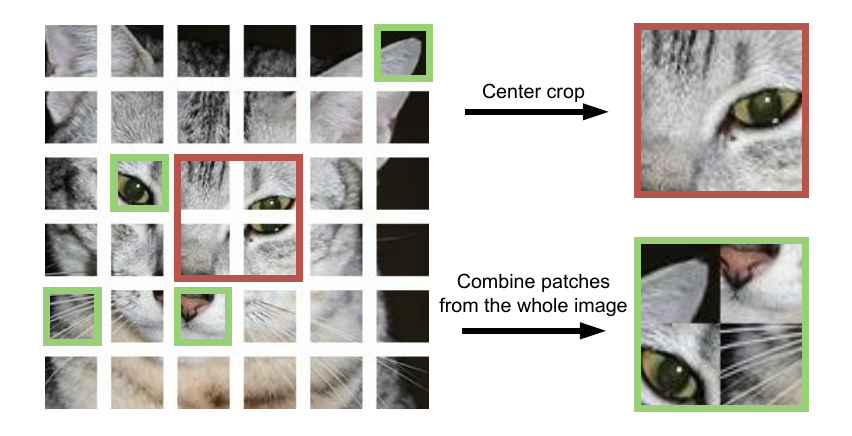}
    \caption{Illustration of the test input processing strategy. In typical methods, a test image is center-cropped before being passed to the detector. Our patch shuffling strategy allows us to select patches from the entire image region, effectively increasing its receptive field.}
    \label{fig:selecting_patches}
    \vspace{-10pt}
\end{figure}

\begin{table}[t]
    \centering
    \resizebox{\linewidth}{!}{
        \begin{tabular}{lcccccc}
\toprule
Benchmark & \multicolumn{2}{c}{SFLD (224+24)} & \multicolumn{2}{c}{SFLD (224+56)} & \multicolumn{2}{c}{SFLD} \\
\cmidrule(lr){2-3} \cmidrule(lr){4-5} \cmidrule(lr){6-7}
& center & full image & center & full image & center & full image \\
\midrule
main benchmark & 98.04 & 98.03 & 98.37 & 98.39 & 98.40 & 98.43 \\
CRN & 94.41 & 96.62 & 94.17 & 97.24 & 91.97 & 95.79 \\
IMLE & 97.55 & 98.65 & 98.12 & 99.23 & 96.92 & 98.64 \\
SITD & 59.36 & 64.82 & 67.71 & 76.66 & 60.38 & 71.90 \\
\bottomrule
\end{tabular}

    }
    \caption{mAP results of the various sizes of test images, comparing two different patch selecting methods. \emph{Center} denotes that the images have been center-cropped to 224×224, while \emph{full image} means that random patches from the full image have been combined to reconstruct a 224×224 image.}
    \label{tab:suppl-fullimage}
\end{table}

\section{Effect of selecting patches from the whole image}
\label{sec:Appendix-fullimagesampling}
\cref{fig:selecting_patches} illustrates the concept of patch extraction of SFLD mentioned in \cref{subsec:SFLD}.
Unlike many alternative detection methodologies, SFLD extracts patches from any position within the input image at the test time.
This approach enhances the detector's receptive field and improves performance for images that have higher resolution than 224×224.
In \cref{tab:suppl-fullimage}, we compare results on benchmarks that have high-resolution images.
We consider different SFLD ensemble options and the location of the selected patch.
The main benchmark consists mostly of 256×256 images, which have little margin with a 224×224 center crop.
Meanwhile, the CRN and IMLE benchmarks have 512×256 images, and the SITD benchmark includes images much larger up to 2,848×4,256 or 4,032×6,030.

We observed that the discrepancy between the two methodologies was minimal when the test image was small.
However, as the image size increased, the performance of the method that solely focused on the center of an image became increasingly constrained.

\section{Image degradation examples}

\begin{figure}[t]
    \centering
    \includegraphics[width=\linewidth]{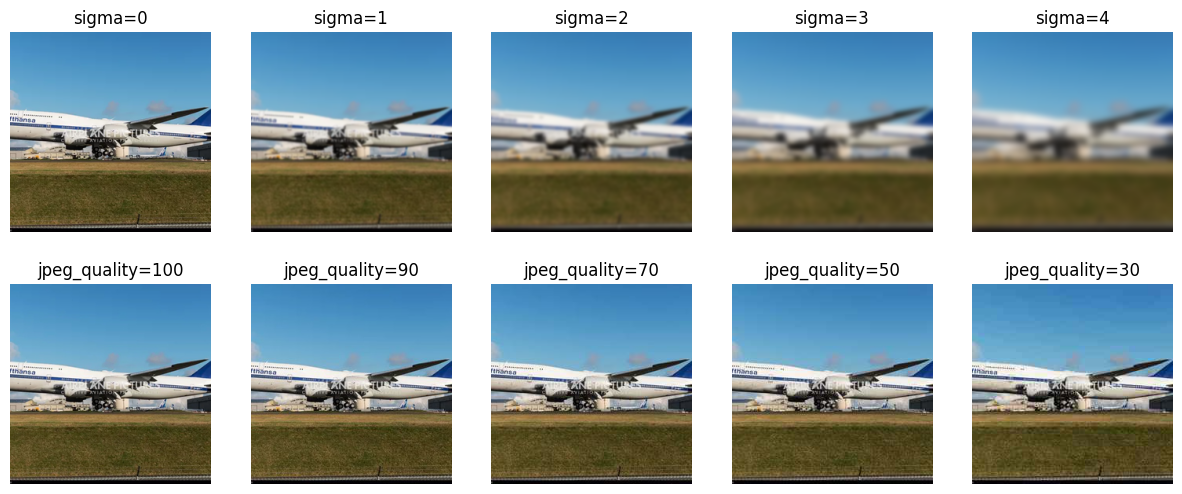}
    \caption{Examples of two image degradation}
    \label{fig:robustness_examples}
\end{figure}

\cref{fig:robustness_examples} shows examples of image gradations.
According to our definition of high- and low-level features, we can consider that the gaussian blur attacks both high- and low-level features in the image, and the JPEG compression attacks on low-level features in the image.

\section{Robustness against image degradation}
\label{sec:supple_robustness}

Since image degradation was not considered during training, it may be useful to examine the changes in output distribution (as shown in \cref{fig:robustness_distribution_jpeg} in supplementary material) to analyze the model's operational tendencies in detail.
\cref{fig:robustness_distribution_jpeg} reveals distinctions between the high-level feature model (UnivFD \cref{fig:robustness_univfd}), low-level feature model (NPR \cref{fig:robustness_npr}), and integrated model.
The distributions of SFLD and UnivFD remain distinguishable, despite a slight decline in discrimination performance.
However, NPR aligns real and generated images into the same distribution.
This behavior arises from the operational mechanism of each model.
NPR primarily focuses on low-level features, resulting in a catastrophic failure to maintain robustness against JPEG compression.
UnivFD demonstrates relative robustness due to its emphasis on high-level features through CLIP visual encoders; however, there is a slight performance penalty because the visual encoder does not completely disregard low-level features.
In contrast, SFLD exhibits robustness against JPEG compression by integrating both high- and low-level features through ensemble/fusion, allowing each to compensate for the information lost in the other.

\section{Effect of patch sizes}

\label{sec:supple_patchsize}
\begin{figure}[t]
    \centering
    \begin{subfigure}[t]{0.48\linewidth}
        \centering
        \includegraphics[width=\linewidth]{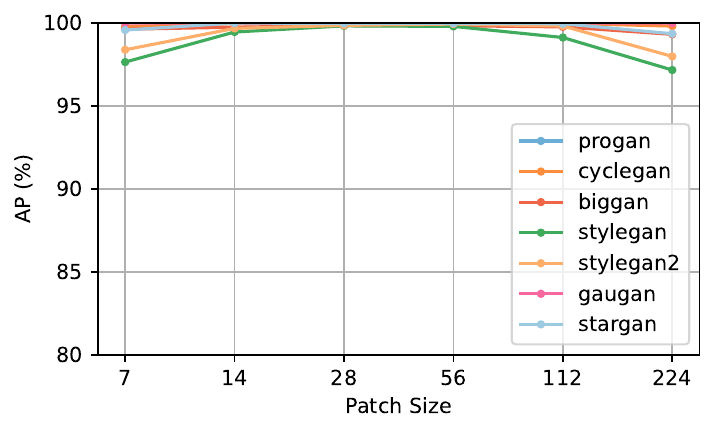}
        \subcaption{GAN-based generators.}
        \label{fig:rebuttal-patchsize-gan}
    \end{subfigure}
    \begin{subfigure}[t]{0.48\linewidth}
        \centering
        \includegraphics[width=\linewidth]{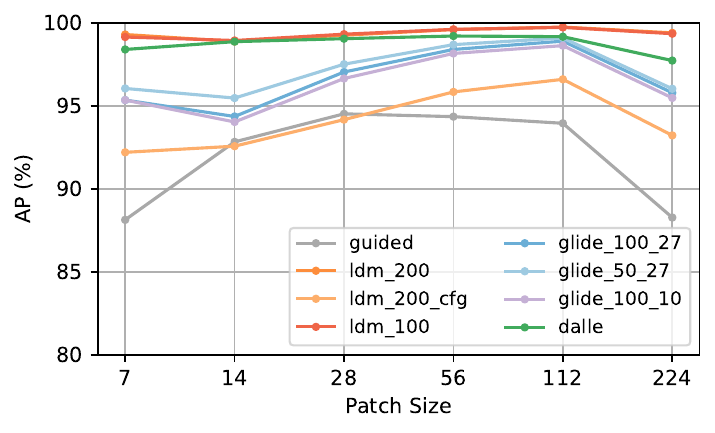}
        \subcaption{Diffusion-based generators.}
        \label{fig:rebuttal-patchsize-diff}
    \end{subfigure}
    \caption{Results of the ensemble models of UnivFD and the patch-shuffled model with each patch size. For 224, it is the same as UnivFD.}
    \label{fig:rebuttal-patchsize}
\end{figure}
To supplement \cref{fig:patch_size} in the main text, we checked the AP for each generator, rather than the average AP on the conventional benchmark.
\cref{fig:rebuttal-patchsize} illustrates that SFLD consistently maintains high performance as long as the patch size is not smaller than the patch size of the image encoder backbone. 
This is because when the shuffling patch size $s_N$ is smaller than the ViT's patch size, the input tokens are affected by patch-shuffling to get an unnatural image patch, resulting in the encoder not properly embedding the visual feature.

\section{Ablation on the pre-trained image encoder}
\label{sec:suppl_pretrained}

\begin{figure}[t]
    \centering
    \begin{subfigure}[t]{0.32\linewidth}
        \centering
        \includegraphics[width=\linewidth]{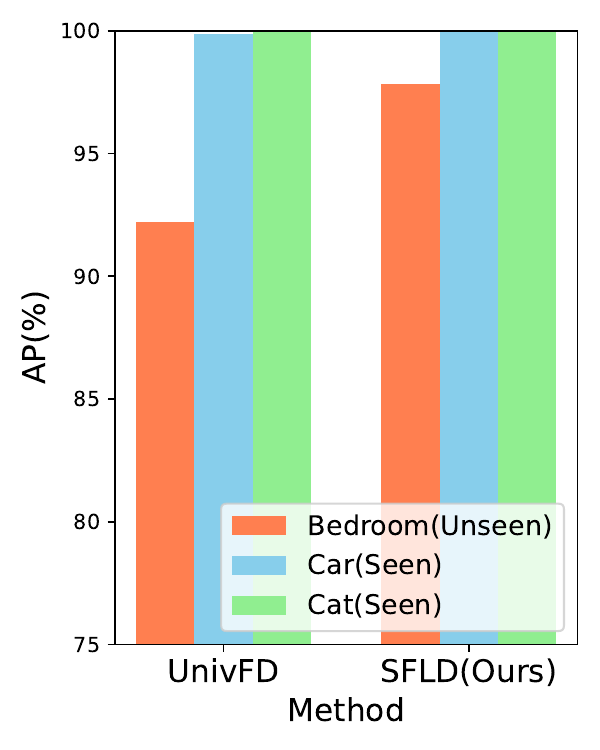}
        \subcaption{CLIP-ViT}
        \label{fig:rebuttal-encoder-clip}
    \end{subfigure}
    \begin{subfigure}[t]{0.32\linewidth}
        \centering
        \includegraphics[width=\linewidth]{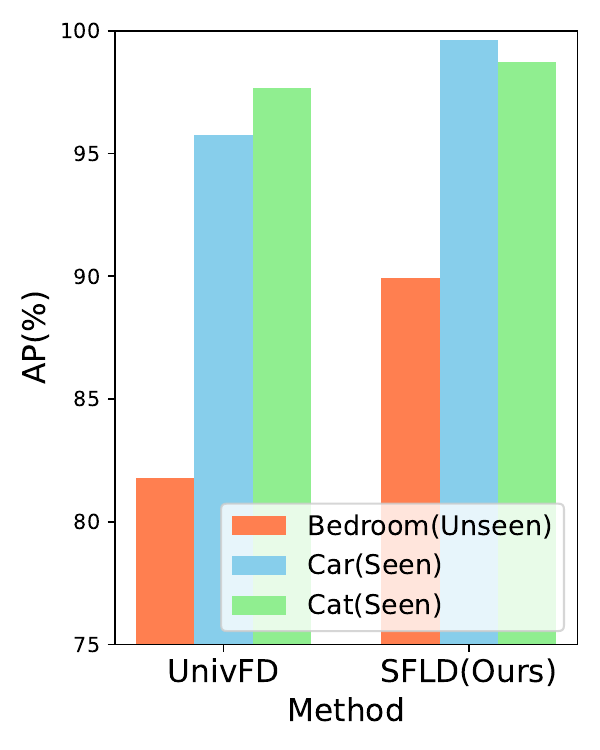}
        \subcaption{DINOv2-ViT}
        \label{fig:rebuttal-encoder-dino}
    \end{subfigure}
    \begin{subfigure}[t]{0.32\linewidth}
        \centering
        \includegraphics[width=\linewidth]{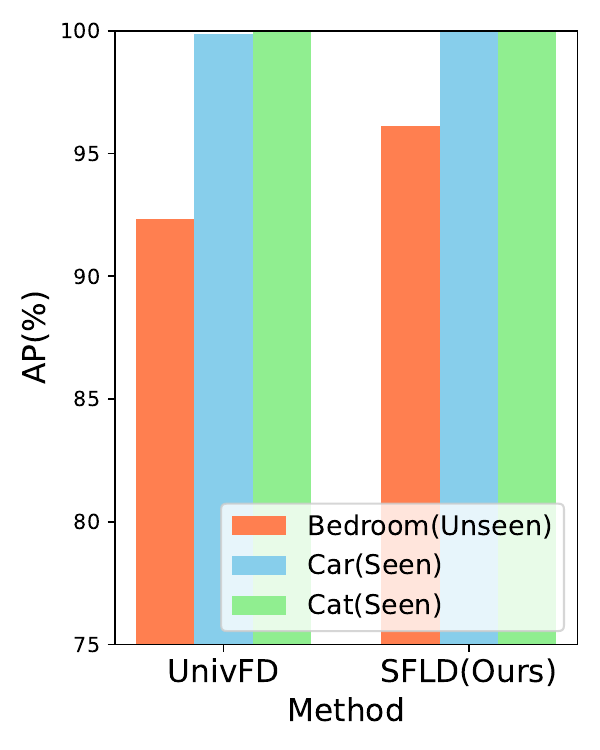}
        \subcaption{OpenCLIP-ViT}
        \label{fig:rebuttal-encoder-openclip}
    \end{subfigure}
    \caption{Class-wise detection results for StyleGAN-\{\emph{bedroom, car, cat}\} class categories reported in AP. \emph{bedroom} class is a novel class that is not in the training set.}
    \label{fig:rebuttal-encoders}
\end{figure}

The pre-trained image encoder is employed to learn the features of the ``real'' class.
According to \cite{ojha2023towards}, directly fine-tuning the encoder makes the detector overfit to a specific generator used in training. This results in low generalization to unseen generators.
Therefore, we utilized the frozen CLIP:ViT-L/14 model following UnivFD.

\cref{tab:rebuttal-encoder} show that our patch shuffling and ensembling strategy improves the performance regardless of the pre-trained backbone.
All models are trained only with real and generated images from ProGAN and tested on the various unseen generated images in conventional benchmark. For ImageNet-ViT, we used ViT-B/16 model, following UnivFD paper \cite{ojha2023towards}. Since its encoders have patch size of 16, we utilized 16 and 32 for patch sizes instead of 28 and 56.
Moreover, note that simply employing different pre-training datasets or strategies -- ImageNet, DINOv2, OpenCLIP -- does not address the content bias problem. (see \cref{fig:rebuttal-encoders})

\begin{figure*}[!h]
    \centering
    \begin{subfigure}{0.32\textwidth}
        \centering
        \includegraphics[width=\linewidth]{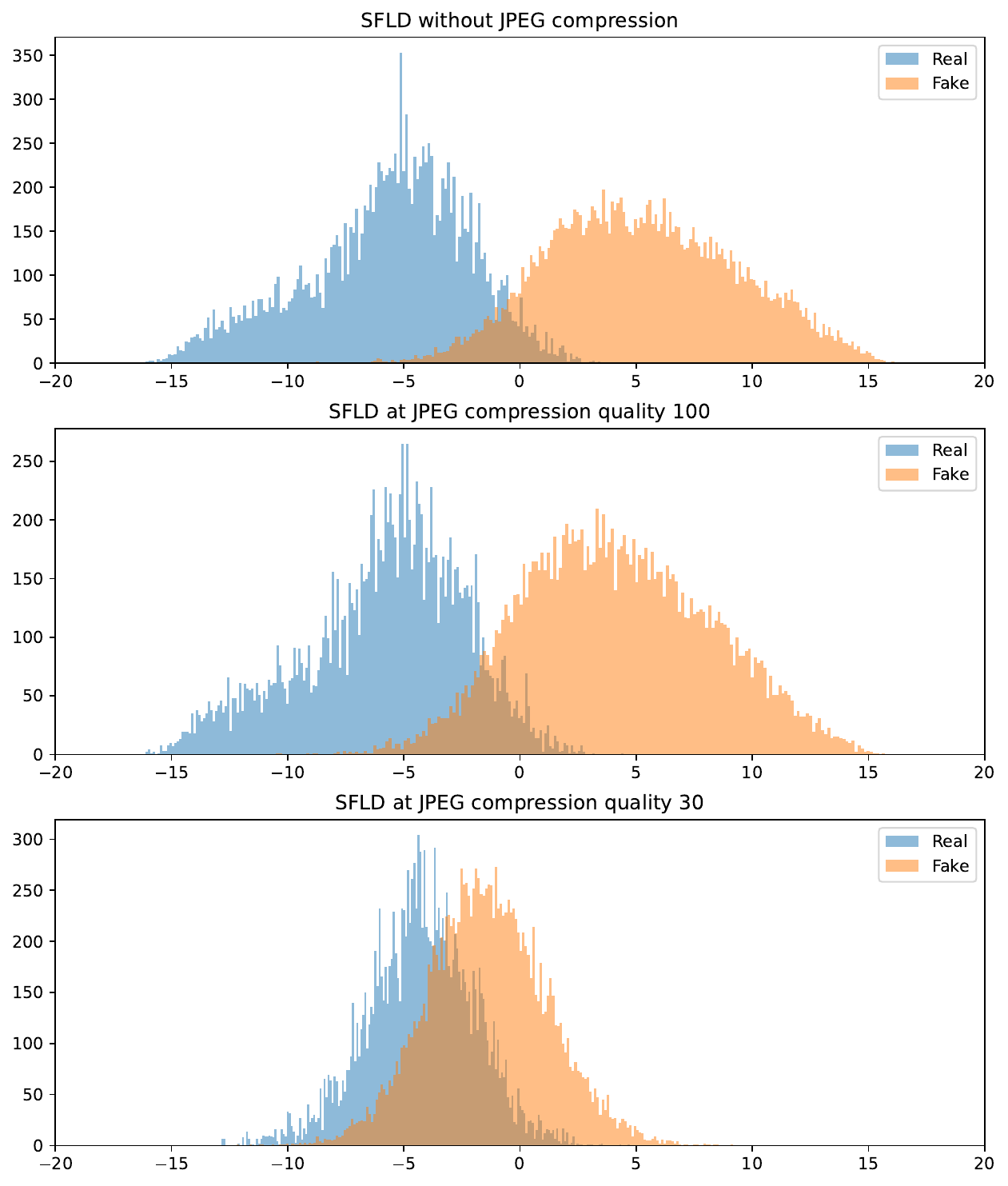}
        \subcaption{The changes of SFLD output distribution}
        \label{fig:robustness_sfld}
    \end{subfigure}
    \begin{subfigure}{0.32\textwidth}
        \centering
        \includegraphics[width=\linewidth]{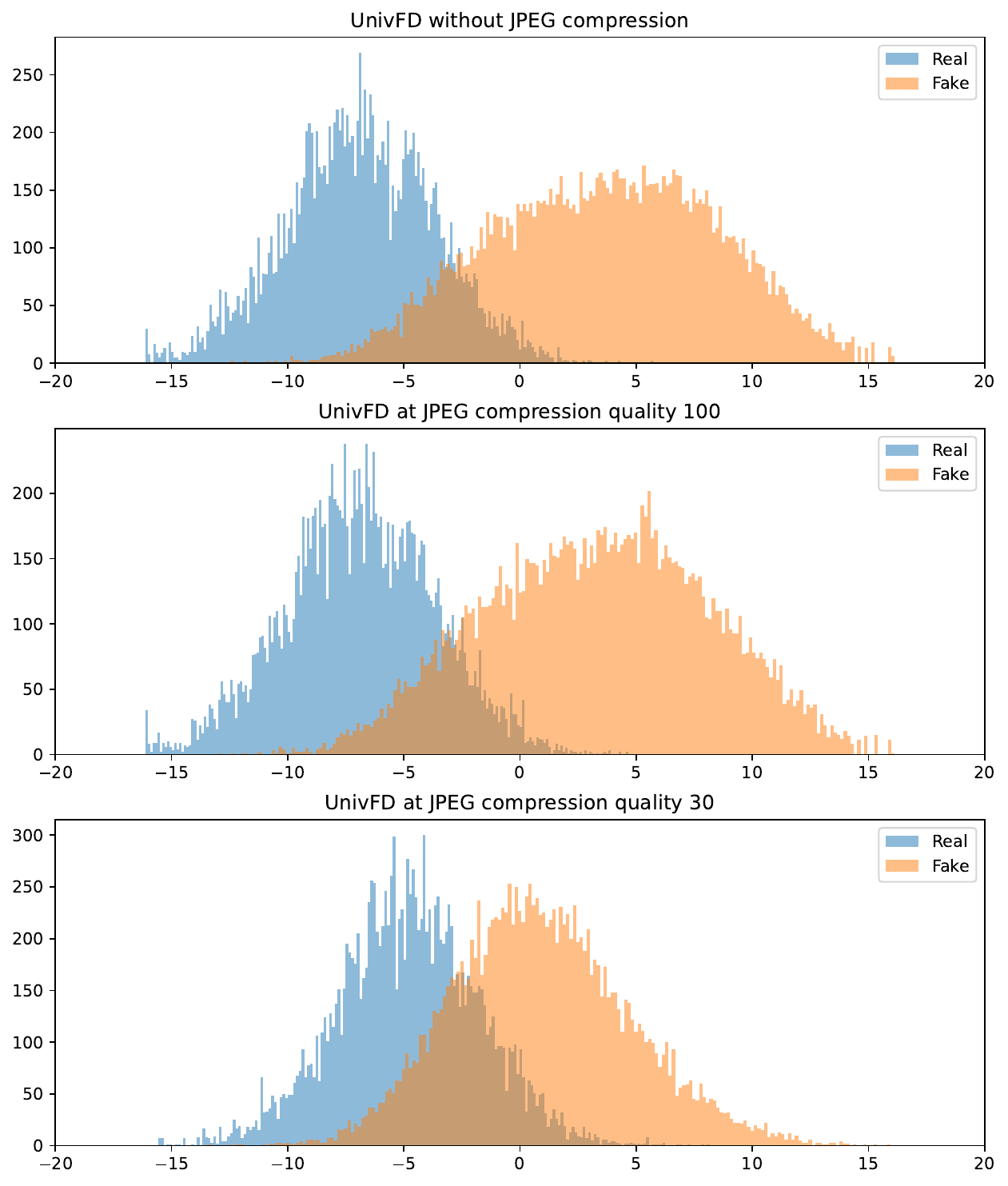}
        \subcaption{The changes of UnivFD output distribution}
        \label{fig:robustness_univfd}
    \end{subfigure}
    \begin{subfigure}{0.32\textwidth}
        \centering
        \includegraphics[width=\linewidth]{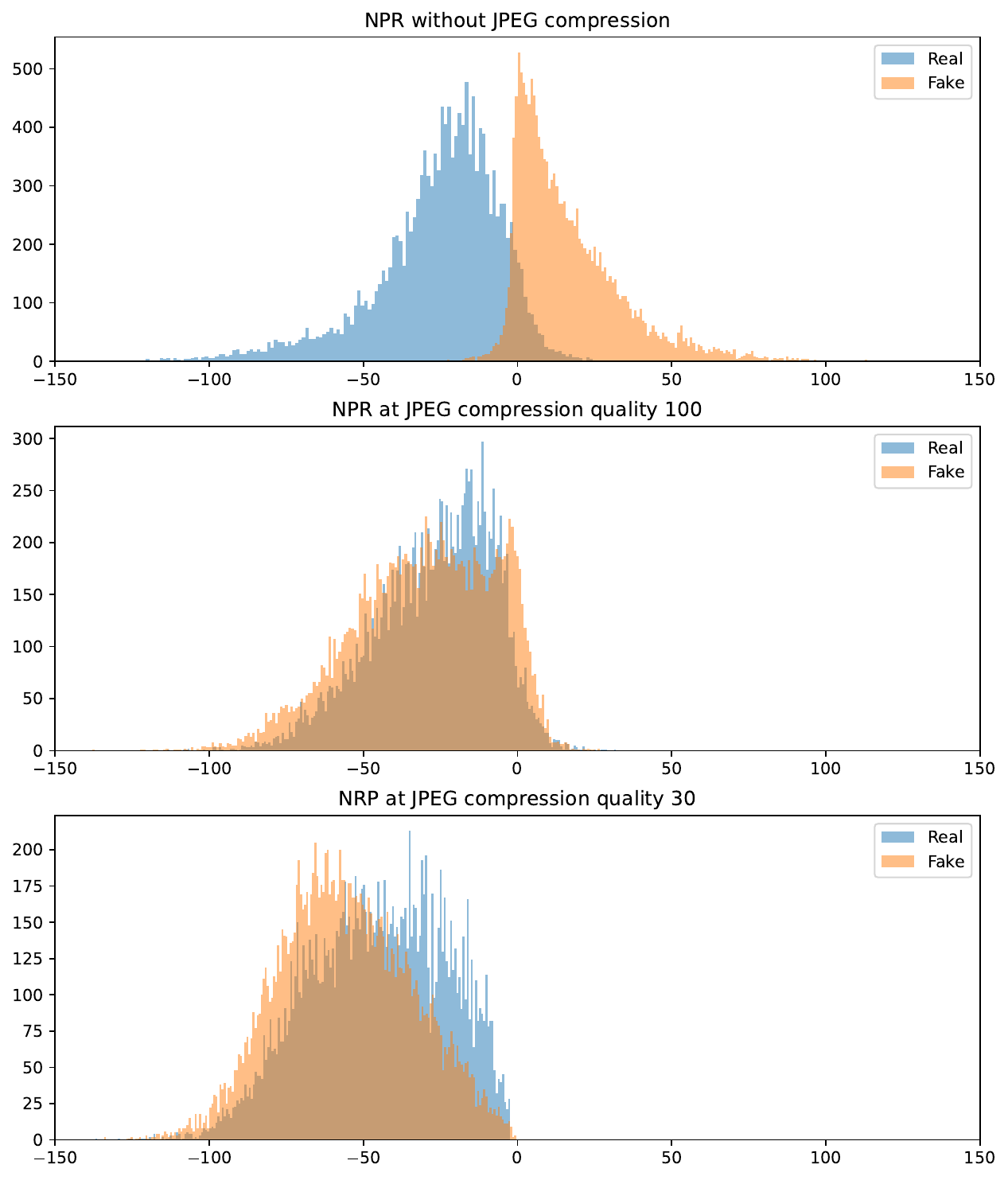}
        \subcaption{The changes of NPR output distribution}
        \label{fig:robustness_npr}
    \end{subfigure}
    \caption{The changes of model output distribution against JPEG compression}
    \label{fig:robustness_distribution_jpeg}
\end{figure*}

\begin{table*}[h]
    \centering
    \resizebox{\textwidth}{!}{
        \definecolor{Gray}{gray}{0.9}
\begin{tabular}{lcc||lcccccc}
    \toprule
    \multirow{2}{*}{\diagbox{Patch sizes}{Pre-training}}  & \multicolumn{2}{c||}{ImageNet-ViT} & \multirow{2}{*}{\diagbox{Patch sizes}{Pre-training}} & \multicolumn{2}{c}{DINOv2-ViT\cite{oquab2023dinov2}} & \multicolumn{2}{c}{OpenCLIP-ViT\cite{ilharco_gabriel_2021_5143773}} & \multicolumn{2}{c}{CLIP-ViT} \\
    \cmidrule(lr){2-3} \cmidrule(lr){5-6} \cmidrule(lr){7-8} \cmidrule(lr){9-10}
    & Acc. & AP & & Acc. & AP & Acc. & AP & Acc. & AP \\
    \midrule
    224 (UnivFD\cite{ojha2023towards})& 62.45 & 69.30 & 224 (UnivFD\cite{ojha2023towards}) & 81.89 & 91.75 & 86.49 & 96.90 & 85.89 & 96.29 \\
    224+16 & 63.88 & 72.23 & 224+28 & 82.88 & 93.42 & 86.50 & 97.59 & 91.94 & 98.03 \\
    224+32 & 63.34 & 71.36 & 224+56 & 82.44 & 93.04 & 86.87 & 97.70 & 92.05 & 98.39 \\
    \rowcolor{Gray} 224+32+16 (ours) & 63.70 & 72.18 & 224+56+28 (ours) & 82.26 & 93.26 & 86.19 & 97.49 & 93.30 & 98.43 \\
    \bottomrule
\end{tabular}
    }
    \caption{Detection accuracy and AP on a conventional benchmark of the proposed patch shuffling and ensembling (SFLD) strategy across various pre-trained encoders. For the ImageNet encoder, ViT-B/16 is used. For the other encoders, ViT-L/14 is used.}
    \label{tab:rebuttal-encoder}

\end{table*}

\section{In-the-wild applications of SFLD}
\label{sec:suppl_inthewild}

We applied our SFLD to in-the-wild AI-generated image detection, especially to a deepfake detection benchmark.
We have already demonstrated performance on a FaceForensics++\cite{roessler2019faceforensicspp} subset, which is a deepfake detection benchmark created using face manipulation software \cite{Deepfakes}. Here, we have added \cref{tab:rebuttal-deepfake} with experiments using Generated Faces in the Wild\cite{borjiGFW} datasets. SFLD shows state-of-the-art performance in detecting real-world deepfakes.

\section{Pseudocode of SFLD}
\label{sec:pseudocode}
See \cref{alg:pseudocode}. 


\begin{algorithm}[h]
    \begin{lstlisting}[language=python]
    """
    Args: 
        image: A test image instance
        n_views: Number of views for random patch shuffle averaging. Defaults to 10. 
        visual_encoder: A CLIP-pretrained ViT-L/14 visual encoder.
    Returns: 
        output: a real/fake score normalized to [0,1] range.
    """

    # prediction from 224x224 unshuffled view
    feature = visual_encoder(image)
    output_224 = classifier_univfd(feature)

    # prediction from 56x56 random shuffled views
    output_56 = []
    for _ in range(n_views): 
        image_shuffled = patch_shuffle(image, size=56)
        feature = visual_encoder(image_shuffled)
        output = classifier_56(feature)
        output_56.append(output)
    output_56 = mean(output_56)

    # prediction from 28x28 random shuffled views
    output_28 = []
    for _ in range(n_views): 
        image_shuffled = patch_shuffle(image, size=28)
        feature = visual_encoder(image_shuffled)
        output = classifier_28(feature)
        output_28.append(output)
    output_28 = mean(output_28)

    # ensemble the logit scores
    output = mean([output_224, output_56, output_28])
    output = output.sigmoid()
    \end{lstlisting}
    \caption{PyTorch-style pseudocode of SFLD}
    \label{alg:pseudocode}
\end{algorithm}

\begin{table}[t]
    \centering
    \resizebox{0.55\linewidth}{!}{
        \definecolor{Gray}{gray}{0.9}

\begin{tabular}{lcc}
\toprule
\multirow[c]{2}{*}{Method} & \multicolumn{2}{c}{GFW\cite{borjiGFW}} \\
\cmidrule(lr){2-3}
 & Acc. & mAP \\
\midrule
NPR\cite{tan2024rethinking} & 53.30 & 47.63 \\
UnivFD\cite{ojha2023towards} & 70.07 & 85.55 \\
SFLD(224+56) & 77.80 & 86.70 \\
\rowcolor{Gray} SFLD & 77.28 & 86.70 \\
\bottomrule
\end{tabular}

    }
    \caption{Performance on the in-the-wild deepfake detection benchmark.}
    \label{tab:rebuttal-deepfake}
\end{table}

\section{Related works}
\label{sec:related-works}

\textbf{AI-generated image detection on specific image generation models} 
Research on distinguishing between synthetic and real images using deep learning models has increased with the development of image generation models. 

Early works were focused on finding the fingerprints in images generated with GANs, which were targeted at high-performing image generation models.
Two major approaches were the use of statistics from the image domain \cite{mccloskey2018detecting, nataraj2019detecting} and the training of CNN-based classifiers.
In particular, in the case of using CNNs, there are two main approaches: focusing on the image domain \cite{mo2018fake, yu2019attributing, tariq2019gan} or the frequency domain \cite{marra2019gans, valle2018tequilagan, Frank}.
Specifically, GAN-generated images have been found to exhibit sharp periodic artifacts in this frequency domain, leading to a variety of applications\cite{ricker2024detectiondiffusionmodeldeepfakes, corvi2023detection, Frank}.

Recently, generative models took a big leap forward with the advent of diffusion models, which called for fake image detection methods that are able to respond to diffusion models.
However, some studies show that existing models trained to detect conventional GANs often fail in images from diffusion models.
For example, periodic artifacts that were clearly visible in GAN were rarely found in diffusion models\cite{ricker2024detectiondiffusionmodeldeepfakes, corvi2023detection}. 
In response, new detection methods optimized for diffusion models have emerged, for example, approaches that use diffusion models to reconstruct test images and evaluate them based on how well they are reconstructed\cite{Wang_2023_ICCV, luo2024lare, zhang2023diffusion}.

\textbf{Generalization of AI-generated image detection}
Recently, the community has shifted its focus towards general AI-generated image detectors that are not specific to GAN or diffusion. In particular, the development of commercially deployed generated models that do not reveal the model structure has increased the demand for such a universal detector.

Apart from existing attempts to learn a specialized feature extractor that simply classifies real/fake in a binary manner, Ojha \etal\cite{ojha2023towards} used the features extracted from a strong vision-language pre-trained encoder that is not trained on a particular AI-generated image. Zhu \etal\cite{zhu2023gendet} combined anomaly detection methods to increase the discrepancy between real and fake image features. 

Furthermore, several studies have concentrated on analyzing pixel-level traces on images inevitably left by the image generators. 
Tan \etal\cite{tan2024rethinking} exploited the artifacts that arise from up-sampling operations, based on the fact that most popular generator architectures include up-sampling operations. 
Chai \etal\cite{chai2020makes} tried to restrict the receptive field to emphasize local texture artifacts.

We design a simple yet powerful general AI-generated image detector that utilizes the feature space of the large pre-trained Vision Language Model.
We apply image reformation to capture not only global semantic artifacts but local texture artifacts from the input images, ensuring detection performance and generalizability on unseen generators.

\end{document}